\documentclass[10pt,twocolumn,letterpaper]{article}

\usepackage[pagenumbers]{cvpr} % To force page numbers, e.g. for an arXiv version

\usepackage{graphicx}
\usepackage{amsmath}
\usepackage{amssymb}
\usepackage{booktabs}
\usepackage{array, makecell} %
\usepackage{pifont}% http://ctan.org/pkg/pifont
\newcommand{\cmark}{\ding{51}}%
\newcommand{\xmark}{\ding{55}}%
\usepackage{multirow}

\usepackage[pagebackref,breaklinks,colorlinks]{hyperref}

\usepackage{xcolor}
\definecolor{darkred}{rgb}{0.7,0,0}
\definecolor{darkgreen}{rgb}{0.1,0.5,0.1}
\definecolor{darkorange}{rgb}{0.6,0.4,0.2}
\definecolor{darkPurple}{rgb}{0.4,0.1,0.4}
\definecolor{lightblue}{rgb}{0.5,0.5,0.75}
\definecolor{blue}{rgb}{0,0,0.75}

\newcommand{\xmarknew}{\textcolor{darkred}{\xmark}}
\newcommand{\cmarknew}{\textcolor{darkgreen}{\cmark}}

\usepackage[capitalize]{cleveref}
\crefname{section}{Sec.}{Secs.}
\Crefname{section}{Section}{Sections}
\Crefname{table}{Table}{Tables}
\crefname{table}{Tab.}{Tabs.}

\def\cvprPaperID{10690} % *** Enter the CVPR Paper ID here
\def\confName{CVPR}
\def\confYear{2023}

\begin{document}

%%%%%%%%% TITLE - PLEASE UPDATE
\title{ Image-Coupled Volume Propagation for Stereo Matching}

\author{Oh-Hun Kwon\\
University of Bonn\\
{\tt\small ohkwon@uni-bonn.de}
% For a paper whose authors are all at the same institution,
% omit the following lines up until the closing ``}''.
% Additional authors and addresses can be added with ``\and'',
% just like the second author.
% To save space, use either the email address or home page, not both
\and
Eduard Zell\\
University of Bonn\\
{\tt\small ezell@uni-bonn.de}
}
\maketitle

%%%%%%%%% ABSTRACT
\begin{abstract}
Several leading methods on public benchmarks for depth-from-stereo rely on memory-demanding 4D cost volumes and computationally intensive 3D convolutions for feature matching. We suggest a new way to process the 4D cost volume where we merge two different concepts in one deeply integrated framework to achieve a symbiotic relationship. A feature matching part is responsible for identifying matching pixels pairs along the baseline while a concurrent image volume part is inspired by depth-from-mono CNNs. However, instead of predicting depth directly from image features, it provides additional context to resolve ambiguities during pixel matching. More technically, the processing of the 4D cost volume is separated into a 2D propagation and a 3D propagation part. Starting from feature maps of the left image, the 2D propagation assists the 3D propagation part of the cost volume at different layers by adding visual features to the geometric context. By combining both parts, we can safely reduce the scale of 3D convolution layers in the matching part without sacrificing accuracy. Experiments demonstrate that our end-to-end trained CNN is ranked 2nd on KITTI2012 and ETH3D benchmarks while being significantly faster than the 1st-ranked method. Furthermore, we notice that the coupling of image and matching-volume improves fine-scale details as demonstrated by our qualitative analysis.

\end{abstract}

%%%%%%%%% BODY TEXT
\section{Introduction}

\begin{figure}[b]
\begin{subfigure}{0.5\linewidth}
  \centering
  \includegraphics[width=.99\linewidth]{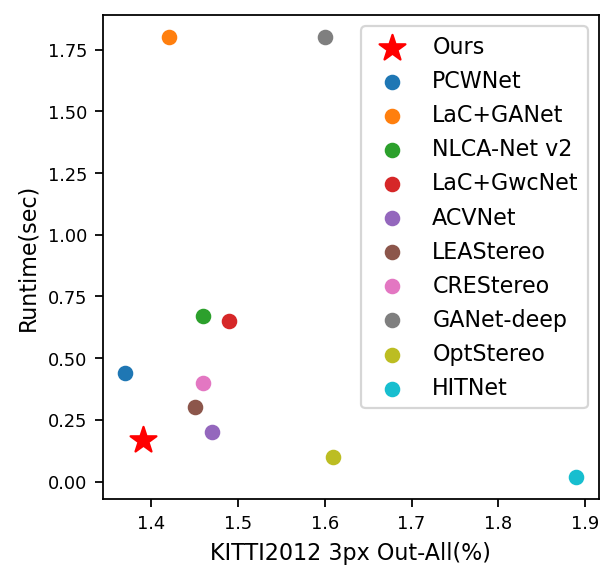}
%   \caption{aa}
  \label{fig:res_kt12}
\end{subfigure}%
\begin{subfigure}{0.49\linewidth}
  \centering
  \includegraphics[width=.99\linewidth]{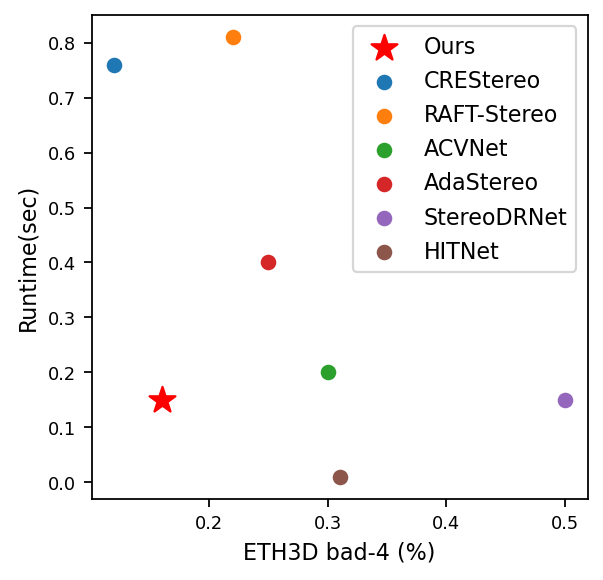}
%   \caption{bb}
  \label{fig:res_e3d}
\end{subfigure}
\caption{
Visual comparison of state-of-the-art methods on KITTI2012 and ETH3D, demonstrating that our method achieves the best trade-off between runtime and accuracy.
}
\label{fig:visual_comparison}
\end{figure}

Depth-from-stereo algorithms are the backbone of a large number of computer vision systems like consumer-grade depth cameras, 3D vision and perception software in robotics, or efficient algorithms for 3D reconstruction and mapping. Even though each domain sets different priorities for accuracy, speed, and robustness, learning-based methods advanced the field over the last years in every aspect. Being such a profound basis, it comes as no surprise that a recent survey \cite{laga2020survey} counted more than 150 learning-based publications for depth from stereo.

A rather popular design choice across many previous learning-based methods is to derive separate feature maps from 2D convolutions for the left and right input image and to construct in the following a 3D or 4D cost-volume for feature matching with the aim to identify globally consistent matching pairs. 
By relying on the completion and smoothing power of 3D convolution, disparity can be estimated even for notoriously difficult cases like uniform or occluded areas or view-dependent shading effects. While being effective, the 4D cost-volume becomes quickly a memory bottleneck. 

To alleviate this limitation we introduce a new variant for 4D cost-volume processing, where we extend the common cost volume aggregation by tightly integrating feature maps of the left image. This addition facilitates the network to estimate disparity from both; the feature matching and the image context. Due to the enhanced input, we can safely reduce the number of features when constructing the cost volume without sacrificing accuracy. Thus despite the addition, our network becomes smaller. Furthermore, the reconstruction of fine-scale details is improved as demonstrated in our qualitative analysis. 
Experiments on established benchmarks demonstrate that computation times and memory use of our method is consistently lower than of comparable methods. At the same time, we achieve comparable or even better scores (Figure \ref{fig:visual_comparison}). 
Our model is ranked 2nd on KITTI2012 and ETH3D benchmarks while being significantly faster than the 1st ranked method.

\section{Related Work}
The original idea of computing depth from a pair of stereo images dates back to the 1970s\cite{marr1976cooperative}. Although, the mathematical principle is clear, solving the pixel matching problem, especially for non-visible or uniform areas as well as fine-scale details remains a challenge. In the following we focus only on the most recent or most similar network designs and refer readers for a more comprehensive overview on learning based methods to the excellent and still very recent survey\cite{laga2020survey}.

As one of the first, MC-CNN\cite{vzbontar2016stereo} introduced feature maps for CNNs to compare image patches. This was followed by GC-Net\cite{kendall2017end}, who established the foundation of probably the most prevailing framework until now, including feature extraction using 2D convolutions, cost volume construction, cost aggregation using 3D convolutions, and disparity regression by the soft argmin function.
While the original cost volume version, concatenating left and right feature maps, remains popular to keep image contexts, several variations to construct the cost volume exist.  
% Cascaded pyramid cost volumes, which are conceptual similarly to image pyramids allow utilizing feature maps in a coarse-to-fine manner\cite{wu2019semantic, yang2019hsm, gu2020cascade, tankovich2021hitnet, shenpcw2022pcw}.
Cascaded pyramid cost volumes, conceptually similar to image pyramids, allow utilizing feature maps in a coarse-to-fine manner\cite{wu2019semantic, yang2019hsm, gu2020cascade, tankovich2021hitnet, shenpcw2022pcw}.
GWCNet\cite{guo2019group} relies on Group-wise correlations, which split the feature vectors into smaller chunks and save correlations of each group in a feature vector within the cost volume. The method remains a popular baseline for other, more recent models\cite{liu2022local, xu2022attention}. Most recent stereo matching models can be classified into three categories: 3D-convolution-based, image-warping-based, and RNN-based models.

% \paragraph*{3D-Convolution-based Models}
The general pipeline for 3D convolution are largely inspired by GC-Net\cite{kendall2017end}.
After constructing the initial cost volume, 3D convolutions filter noisy voxels or smoothen sparsely matched parts based on the 3D geometric context.
HSM-Net\cite{yang2019hsm} propagates cascaded pyramid cost volumes in a coarse-to-fine manner, aiming for stereo matching with high-resolution images.
StereoDRNet\cite{chabra2019stereodrnet} employs 3D atrous convolutions with different dilation, whose results are concatenated into a single cost volume.
Our model adopts atrous convolutions but propagates 2D feature maps and 3D cost volumes simultaneously before merging.
Given the general pipeline with 2D convolution layers for feature extraction and 3D convolution layers for cost aggregation, LEAStereo\cite{cheng2020hierarchical} applied Neural Architecture Search(NAS) technique to automatically identify the most efficient network structure instead of designing the neural network manually.
Based on GWCNet\cite{guo2019group} with the stacked hourglass structure\cite{newell2016stacked}, ACVNet\cite{xu2022attention} proposed attention concatenation volume, intending to filter out redundant information of the initialized cost volume and make the model more efficient.
Similarly, we demonstrate that the size of 3D convolution layers can be down-scaled by leveraging a concurrent 2D image pipeline. 

% \paragraph*{Image-Warping-based Models}
Image warping shifts pixels or feature maps of the right image based on the estimated disparity with the aim to achieve overlapping image pairs.  % except for the invisible parts from the right image.
By comparing the left and the warped right feature map or 2D image, 2D convolution predict the current errors, which is subsequently refined\cite{pang2017cascade}.
HITNet\cite{tankovich2021hitnet} adopts the slanted-plane principle\cite{tankovich2018sos} enhancing the quality during up-sampling through a coarse-to-fine design. % with the estimated gradient.
Alternatively, a full-cost volume is transformed into a thinner volume, by limiting the disparity of each pixel, which in the end is equivalent to image warping\cite{mao2021uasnet, shenpcw2022pcw}. %of the prediction
While alleviating the computation by limiting the disparity range, 
models relying on image warping neglect the 3D geometric context of the full cost volume.

% \paragraph*{RNN-based Models}
RAFT\cite{teed2020raft} was originally introduced with remarkable results for optical flow and was successfully adapted to stereo matching problems\cite{lipson2021raft}.
Starting from a cost volume RAFT iteratively optimizes disparity by searching the cost volume with Recurrent Neural Network (RNN).
CREStereo is a cascaded version of RAFT with adaptive group correlation layers achieving state-of-the-art results\cite{li2022practical}.
ORStereo\cite{hu2021orstereo} adopts the RAFT mechanism for stereo matching for image resolutions of up to 4K. %, where image warping is collaboratively used.
Although the iterative design of RNN-based methods offers the potential of handling larger images, it requires in general longer computation times. %produce significantly smooth results through many iterations but

Image-warping and RNN-based models process the left image feature map separately through their pipeline and only few 3D-convolution methods incorporate an independent path for the 2D feature map.
Stereonet\cite{khamis2018stereonet} and Activestereonet\cite{zhang2018activestereonet} utilize a 2D feature map to refine regressed disparity at the ultimate stage, which is not included into the 3D cost volume to reduce inference. 
%where fundamental inference mainly performs using 3D geometric context.
Inspired by semi-global matching\cite{hirschmuller2007stereo}, GA-Net\cite{Zhang2019GANet} proposes learnable, semi-global aggregation, which integrates feature maps from the left image into the cost volume. However, their row and column wise operation within the feature map slows down the overall computation.
Meanwhile, our efficient 2D/3D fusion lightly affects the computation resource and is more frequently connected through the whole network to refine the geometric context of the cost volume.

\begin{figure*}[t]
  \begin{center}
  \includegraphics[width=1\linewidth]{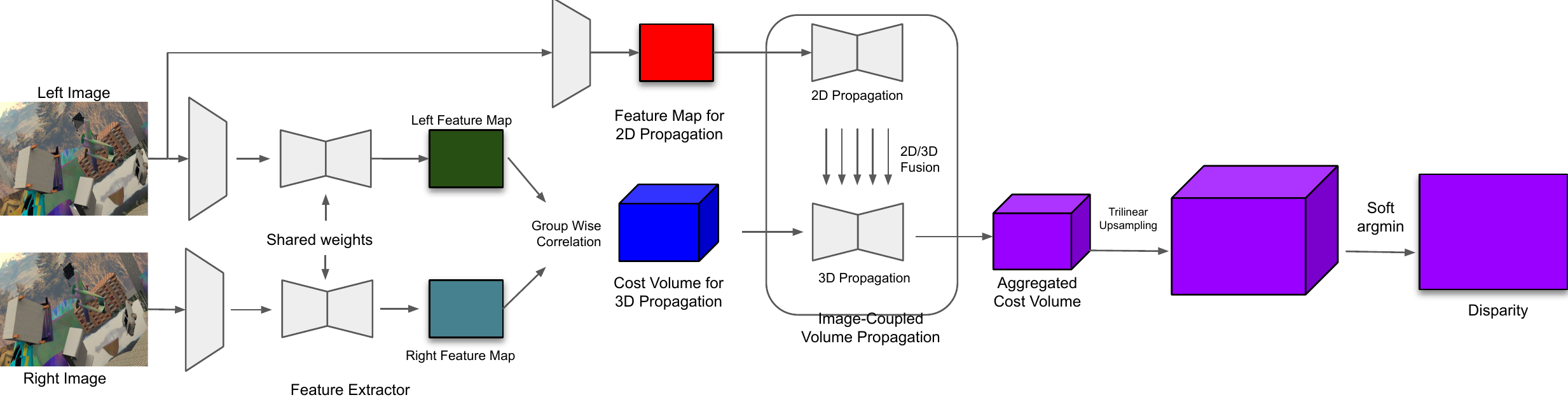}
\end{center}
\caption{
The full architecture of our model, consisting of feature extraction, cost volume construction, cost aggregation, and disparity regression.
}
\label{fig:wholemodel}
\end{figure*}

\section{Network Architecture}
% % % % % % % % % % % % % % % % % % % % % % % % % % % % % % % % 
The general architecture of our model follows recent stereo matching methods\cite{kendall2017end, cheng2020hierarchical, guo2019group, xu2022attention}. It consists of four building blocks: feature extraction, cost volume construction, cost aggregation, and disparity regression which we first describe briefly (section \ref{sec:high_level_arch}) and focus in detail on our main contributions for the cost-volume in later sections. 

%\edc{General comment: I would add a few hints (mostly one sentense or less is enough) how the benefits (memory, computation time) are achieved compared to previous work.}
%\note{The basedline model is smaller. I used UNet while others use tripled-stack Hourglass. But I don't think it is not the main contribution. It's just a smaller model}

\subsection{High-Level Architecture}\label{sec:high_level_arch}
\paragraph*{Feature Extractor: }
% \edc{$c_f$ is not defined.}
Given a pair of left and right color images $\mathbf{I}_l, \mathbf{I}_r \in \mathbb{R} ^ {3 \times H \times W}$, the feature extractor, a UNet-like CNN\cite{ronneberger2015u}, produces the feature maps $\mathrm{f}_l, \mathrm{f}_r \in \mathbb{R}^{c_f \times h_0 \times w_0}$ where $h_0 = \left \lfloor H / 3 \right \rfloor$ and $w_0 = \left \lfloor W / 3 \right \rfloor$ and $c_f$ is the dimension of feature vectors. The feature extractor starts with three layers of $3\times3$ 2D convolutions and outputs  16, 32, and 32 channels with strides of 1, 3, and 1. The original images are first downscaled to a third of the input resolution, followed by a 3-level UNet. The UNet design was preferred because it turned out more efficient and led to higher accuracy during early development.
% \note{reminder} A high dimensional feature vector is required for the input of the cost volume.
% \note{reminder} A high dimensional feature vector is required for the input of the cost volume.
%\note{in my case yes. And less gpu memory and higher resolution 1/3 vs 1/4} \note{for other method cases, the stack hourgalss is used on 4D cost aggregation, not for feature extraction }
%\edc{I would leave this out for now: which will be constructed by group-wise correlation\cite{guo2019group}}. 
% To satisfy the requirements we use an asymmetric encoder-decoder structure and increase the number of output channels of the feature extractor.
In preparation of group-wise correlations (GWC) and in order to achieve high dimensional vectors for the feature maps, we use an asymmetric encoder-decoder structure and increase the number of output channels of the feature extractor.
The output dimensions are 32, 64, and 128 for each encoder layer and 128 for all decoder layers.

\paragraph*{Cost Volume Construction:}
% To operate group-wise correlations, each feature map \edc{should this not be ${f}^{g}$ or later only ${f}$} $\mathrm{f} \in \mathbb{R}^{c_f \times h_0 \times w_0}$ is split along the channel dimension into $c_0$ number of feature maps $\left\{ \mathrm{f}^{1}, \mathrm{f}^{2}, ..., \mathrm{f}^{c_0}\right\}$ where each $\mathrm{f}^{g}$ has $c_f/c_0$ number of channels. 
To profit from higher accuracy of group-wise correlations (GWC), each feature map $\mathrm{f} \in \mathbb{R}^{c_f \times h_0 \times w_0}$ is split along the channel dimension into $c_0$ number of feature maps $\left\{ \mathrm{f}^{g} \right\}_{g=1,..,c_0}$ where the number of channels is $c_f/c_0$ for each $\mathrm{f}^{g}$. 
Each value of the cost volume $v_0 \in \mathbb{R} ^ {c_0 \times d_0 \times h_0 \times w_0}$ is initialized by 
\begin{align}
    \mathbf{v}_0\left(g, d, i, j\right)=\frac{1}{c_f / c_0}\left\langle\mathrm{f}_l^g(i, j), \mathrm{f}_r^g(i, j-d)\right\rangle
\end{align}
where $d_0 = \left \lfloor D / 3 \right \rfloor$, $D$ is the maximum disparity and $i,j$ are the pixel idices.

\paragraph*{Cost Volume Aggregation}
%\edc{Note to myself: rename image feature/image map in cost volume to reduce confusion.}
The cost aggregation module consists of two parallel, one-directionally connected UNet structures. The 2D propagation part evolves an additional feature map $\mathbf{f}_{0}$ of the left image, that is similar by design to the feature extractor, but independent (see Figure \ref{fig:wholemodel}). To build $\mathbf{f}_{0} \in \mathbb{R}^{c_0 \times h_0 \times w_0}$, the left image $\mathbf{I}_l$ passes through three layers of $3\times3$ 2D convolutions to fit the size of the feature map to $\mathbf{v}_0$ for the 2D/3D fusion.
The 3D propagation part convolves the 4D cost volume $\mathbf{v}_0$ 
while incorporating the feature maps $\mathbf{f}_{i}$ from the input images to produce the jointly evolved 4D cost volume $\mathbf{v}_{end} \in \mathbb{R} ^ {d_0 \times h_0 \times w_0}$.

\begin{figure*}[t]
  \begin{center}
  \includegraphics[width=1\linewidth]{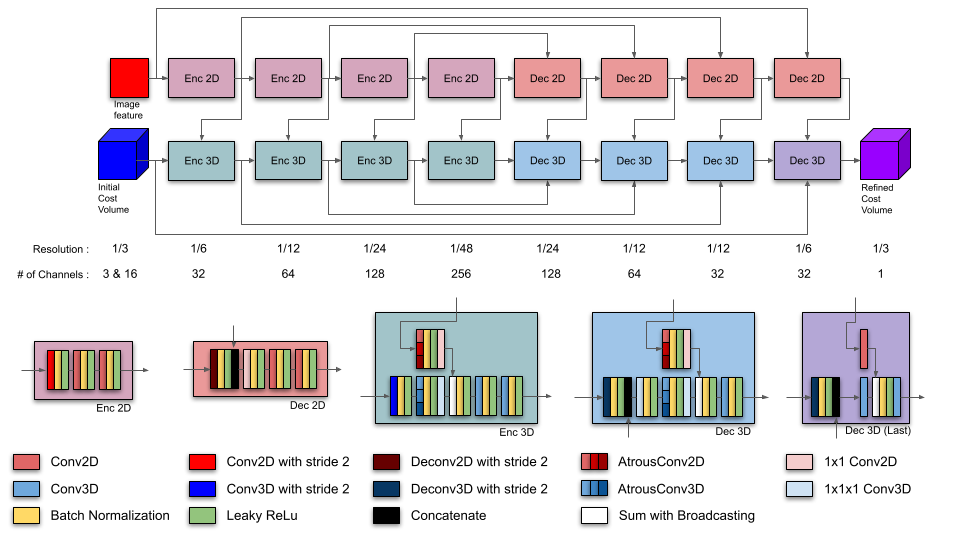}
\end{center}
\caption{
Image-coupled Volume Propagation consists of two paths for the 2D feature maps $\mathbf{f}_i$ and the 3D cost volume $\mathbf{v}_i$ each of which has UNet-like structures. Each output from a 2D propagation is connected to the corresponding 3D propagation with atrous convolution and broadcasting sum.
}
\label{fig:model}
\end{figure*}

% The cost aggregation module has two parallel paths of UNet structures to evolve each of the left feature map $\mathbf{f}_{0}$ and the cost volume $\mathbf{v}_0$. 
At each encoder and decoder level, the 3D UNet for $\mathbf{v}_i$ has an input connection from the corresponding level of the 2D UNet (Figure \ref{fig:model}) thus each feature map $\mathbf{f}$ and cost volume $\mathbf{v}$ are evolving at level $n$ as:
%\note{Original. At each level of encoder and decoder, the 3D UNet for $\mathbf{v}_i$ has an input connection from the corresponding level of the 2D UNet evolving $\mathbf{f}_i$ as}
\begin{align}
\begin{split}\label{eq:fenc}
    \mathbf{f}^\mathrm{Enc}_{n} = \footnotesize\text{Enc}^{2d}\left(\mathbf{f}^\mathrm{Enc}_{n-1}\right)
\end{split}\\
\begin{split}\label{eq:fdec}
    \mathbf{f}^{\mathrm{Dec}}_{n} = \footnotesize\text{Dec}^{2d}\left(\mathbf{f}^{\mathrm{Dec}}_{n+1}, \mathbf{f}^\mathrm{Enc}_{n}\right)
\end{split}\\
\begin{split}\label{eq:venc}
    \mathbf{v}^\mathrm{Enc}_{n} = \footnotesize\text{Enc}^{3d}\left(\mathbf{v}^\mathrm{Enc}_{n-1}, \mathbf{f}^\mathrm{Enc}_{n}\right)
\end{split}\\
\begin{split}\label{eq:vdec}
    \mathbf{v}^{\mathrm{Dec}}_{n} = \footnotesize\text{Dec}^{2d}\left(\mathbf{v}^{\mathrm{Dec}}_{n+1}, \mathbf{v}^\mathrm{Enc}_{n}, \mathbf{f}^\mathrm{Dec}_{n}\right)
\end{split}
\end{align}
where $\mathbf{f}^\mathrm{Enc}_{0} = \mathbf{f}_{0}$ and $\mathbf{v}^\mathrm{Enc}_{0} = \mathbf{v}_0$.
More details on the cost volume aggregation are described in sections \ref{sec:2d_3d_fusion} and \ref{sec:volume_propagation}.

\paragraph*{Disparity Regression}
To recover the original resolution of the input images, the cost volume $\mathbf{v}^{\mathrm{Dec}}_{0}$ is up-sampled by tri-linear interpolation to the cost volume $\mathbf{V} \in \mathbb{R} ^ {D \times H \times W}$. 
The disparity $\mathbf{d}$ is estimated for each pixel with the soft argmax function\cite{kendall2017end},
\begin{align}
    \mathbf{d}\left(i, j\right)=\sum_{d=0}^{D-1} \frac{ e^{V(d,i,j)}}{\sum_{k} e^{V(k,i,j)}} d .
\end{align}

\paragraph*{Loss Function}
During all trainings, we use smooth L1 loss in an end-to-end manner, where the loss is defined as:
%$\mathcal{L}$ where %$=\operatorname{smooth}_{L_1}\left(\mathbf{d} -\mathbf{d}_{\mathrm{gt}}\right)$, 
%\edc{not sure, but prevents introducing $x$}
\begin{align}
    \mathcal{L}=\left\{\begin{array}{lc}
    0.5 \left(\mathbf{d} -\mathbf{d}_{\mathrm{gt}}\right)^2, & |\mathbf{d} -\mathbf{d}_{\mathrm{gt}}|<1 \\
    |\mathbf{d} -\mathbf{d}_{\mathrm{gt}}|-0.5, & \text { otherwise }
    \end{array}\right.
\end{align}
Experiments with other and more sophisticated loss functions did not outperform the smooth L1 loss.

\begin{figure}[t]
  \begin{center}
  \includegraphics[width=1\linewidth]{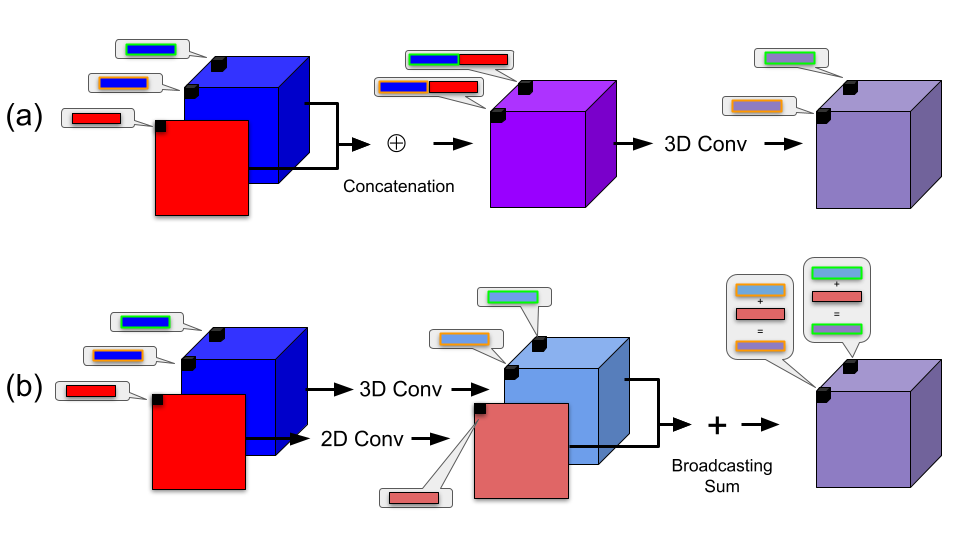}
\end{center}
\caption{
Illustration of 2D-3D fusion by broadcasting sum. The broadcasting sum results in the same computation as the 2D-3D feature concatenation while being more efficient in computation and memory consumption.
}
\label{fig:fusion}
\end{figure}

% % % % % % % % % % % % % % % % % % % % % % % % % % % % % % % % 
\subsection{Cost Volume - 2D/3D Fusion}\label{sec:2d_3d_fusion}

% \edc{ist't it a 4D and not a 3D cost volume?}
% ----------------------------------

% ----------------------------------
% \note{actually true but 3D/4D fusion sounds very misleading. since it is fusion of feature maps in image space and cost volume in 3D space I would say it is better}
% \note{also there are 2D Unet and 3D Unet. So it would make sense for pipeline, methods, model names whild we should be careful on using 2D/3D on feature maps and cost volume}
%was: \note{in a 2D image space}\note{in a 3D space}
A straightforward way to incorporate the feature map of the 2D propagation part into the 3D propagation part is to concatenate the feature maps at each sliced plane of the cost volume along the disparity dimension.
% In this case, a voxel of the cost volume will embed both the geometric context from the input cost volume and the image feature vector at the corresponding pixel of the feature map.
In this case, a voxel of the cost volume will embed a vector of both the geometric context from the input cost volume and the corresponding pixel of the feature map.
If we concatenate the cost volume $\mathbf{v}$ and feature map $\mathbf{f}$ into the volume $\mathbf{v}'$, a voxel can be formulated as
% \edc{changed v and f to not bold, only the maps/vectors are bold}
% \note{they are actually the feature maps and cost volumes}
% \edc{My assumption is that a map has several entries (thus bold), but a voxel entry is a scalar (normal font). Due to the index variables $i$ and $j$, $v(c,d,i,j)$ is a scalar of map $\mathbf{v}$}
% \begin{align}
% \begin{split}
%     v'\left(c, d, i, j\right) = \left\{\begin{array}{lc}
%     v\left(c, d, i, j\right), & 1 \leq c \leq C_v \\
%     f\left(c - C_v, i, j\right), & C_v+1 \leq c \leq C_v + C_f
%     \end{array}\right.
% \end{split}
% \end{align}
\begin{align}
\begin{split}
    \mathbf{v}'\left(c, d, i, j\right) = \left\{\begin{array}{lc}
    \mathbf{v}\left(c, d, i, j\right), & 1 \leq c \leq C_v \\
    \mathbf{f}\left(c - C_v, i, j\right), & C_v+1 \leq c \leq C_v + C_f
    \end{array}\right.
\end{split}
\end{align}
where $C_v$ and $C_f$ are the dimensions of the vectors $\mathbf{v}$ and $\mathbf{f}$. 

% \note{
% $\mathbf{v}' \in \mathbb{R} ^ {(C_v + C_f) \times D \times H \times W}$
% $\mathbf{v}'\left(c, d, i, j\right)$ is a scalar
% $\mathbf{v} \in \mathbb{R} ^ {C_v \times D \times H \times W} $
% $\mathbf{v}\left(c, d, i, j\right)$ is a scalar
% $\mathbf{f} \in \mathbb{R} ^ {C_f \times H \times W}$
% $\mathbf{f}\left(c - C_v, i, j\right)$ ia a scalar
% }

If 3D convolution is applied on the concatenated cost volume $\mathbf{v}'$, the operation is equivalent to the broadcasting sum of the feature map and the cost volume, each of which is processed by a 2D and 3D convolution, as shown in
%\edc{changed v and f to not bold, only the maps/vectors are bold. $w$ and $W$ are not defined, only $w_o$ and $W$ are defined.}
%\note{Ok. Definition of W are duplicated. Here, it was weights of the convolutions. It should be redefined. $\mathbf{W}$}
% \begin{align}
% \begin{split}    &\footnotesize\text{Conv}^{3d}_{k,s}\left(v'; \mathbf{W}_{v'} \right)\left(c, d, i, j\right)
% \\ = &\sum_{\bar{c}} \sum_{\bar{d}} \sum_{\bar{i}} \sum_{\bar{j}} \mathbf{w}_{\bar{c},\bar{d},\bar{i},\bar{j}}v'\left(c+\bar{c}, d+\bar{d}, i+\bar{i}, j+\bar{j}\right) 
% \\ =&\sum_{\bar{c}=1}^{C_v} \sum_{\bar{d}} \sum_{\bar{i}} \sum_{\bar{j}} \mathbf{w}_{\bar{c},\bar{d},\bar{i},\bar{j}}v\left(c+\bar{c}, d+\bar{d}, i+\bar{i}, j+\bar{j}\right) + 
% \\ & \sum_{\bar{c}=1}^{C_f} \sum_{\bar{i}} \sum_{\bar{j}} \left(\sum_{\bar{d}} \mathbf{w}_{\bar{c}+C_v,\bar{d},\bar{i},\bar{j}} \right)\mathbf{f}\left(c - C_v +\bar{c}, i+\bar{i}, j+\bar{j}\right) 
% \\ =&\footnotesize\text{Conv}^{3d}_{k,s}\left(\mathbf{v}; \mathbf{W}_v \right)\left(c, d, i, j\right)
% + \footnotesize\text{Conv}^{2d}_{k,s}\left(f; \mathbf{W}_f \right)\left(c, i, j\right).
% \end{split}
% \end{align}
\begin{align}
\begin{split}    &\footnotesize\text{Conv}^{3d}_{k,s}\left(\mathbf{v}'; \mathbf{W}_{v'} \right)\left(c, d, i, j\right)
\\ = &\sum_{\bar{c}} \sum_{\bar{d}} \sum_{\bar{i}} \sum_{\bar{j}} \mathbf{w}_{\bar{c},\bar{d},\bar{i},\bar{j}}\mathbf{v}'\left(c+\bar{c}, d+\bar{d}, i+\bar{i}, j+\bar{j}\right) 
\\ =&\sum_{\bar{c}=1}^{C_v} \sum_{\bar{d}} \sum_{\bar{i}} \sum_{\bar{j}} \mathbf{w}_{\bar{c},\bar{d},\bar{i},\bar{j}}\mathbf{v}\left(c+\bar{c}, d+\bar{d}, i+\bar{i}, j+\bar{j}\right) + 
\\ & \sum_{\bar{c}=1}^{C_f} \sum_{\bar{i}} \sum_{\bar{j}} \left(\sum_{\bar{d}} \mathbf{w}_{\bar{c}+C_v,\bar{d},\bar{i},\bar{j}} \right)\mathbf{f}\left(c - C_v +\bar{c}, i+\bar{i}, j+\bar{j}\right) 
\\ =&\footnotesize\text{Conv}^{3d}_{k,s}\left(\mathbf{v}; \mathbf{W}_v \right)\left(c, d, i, j\right)
+ \footnotesize\text{Conv}^{2d}_{k,s}\left(\mathbf{f}; \mathbf{W}_f \right)\left(c, i, j\right). \; ,
\end{split}
\end{align}
where $\mathbf{W}_{v'}$, $\mathbf{W}_v$, and $\mathbf{W}_f$ are the corresponding sets of the convolution weights $\mathbf{w}$.

While the concatenation produces a cost volume $\mathbf{v}' \in \mathbb{R} ^ {\left(C_v + C_f\right) \times D \times H \times W}$, the broadcasting sum intermediately has a cost volume $\mathbf{v} \in \mathbb{R} ^ {C_v \times D \times H \times W}$ and a feature map $\mathbf{f} \in \mathbb{R} ^ {C_f \times H \times W}$ resulting in less weight numbers, GPU memory consumption, and computation use.
Figure \ref{fig:fusion} illustrates the operations of the concatenation and the broadcasting sum to fuse a feature map and a cost volume.
To preserve equivalence with the concatenating method, it is required to compute the broadcasting sum before normalization and activation layers.

%\edc{highlight atrous more to justify comparison in ablation study}
% In addition to broadcasting sum, we adopt atrous convolutions\cite{florian2017rethinking} on both 2D and 3D outputs to involve larger receptive fields.
%\edc{Are atrous convolution present in feature map generation? If yes, we should add this detail. If not we should highlight it here.}
%\note{No it is only on cost aggregation for our model. the reference used it on both feature extraction and cost aggregation }
In addition to the broadcasting sum, we adopt atrous convolutions, similar to \cite{chabra2019stereodrnet} but in our case on both 2D and 3D inputs, to obtain larger receptive fields and to improve accuracy. 
Each atrous block is the concatenation of three convolutions with different dilations as
\begin{align}
\begin{split}
& \gamma^{2d}\left(x\right) =  \left\{{\footnotesize\text{Conv}}^{2d}_{_{|1,1|}}\left(x\right), {\footnotesize\text{Conv}}^{2d}_{_{|2,2|}}\left(x\right), {\footnotesize\text{Conv}}^{2d}_{_{|3,3|}}\left(x\right)\right\}
\end{split}\\
\begin{split}
& \gamma^{3d}\left(x\right) =  \left\{{\footnotesize\text{Conv}}^{3d}_{_{|1,1,1|}}\left(x\right), {\footnotesize\text{Conv}}^{3d}_{_{|1,2,2|}}\left(x\right), {\footnotesize\text{Conv}}^{3d}_{_{|1,3,3|}}\left(x\right)\right\}
\end{split}
\end{align}
where ${\footnotesize\text{Conv}}^{Xd}_{_{|d|}}$ is X-dimensional convolution with the kernel size 3, stride 1, and dilation $d$.

For the following formulations, we denote the sequential operations of convolution, batch normalization\cite{ioffe2015batch}, and Leaky ReLU\cite{maas2013rectifier} as
\begin{align}
    \Phi^{Xd}_{k,s}\left(x\right) = \footnotesize\text{LeakyReLU}_{0.1}\left(\footnotesize\text{BatchNorm}\left({\footnotesize\text{Conv}}^{Xd}_{k,s}\left(x\right)\right)\right)
\end{align}
where ${\footnotesize\text{Conv}}^{Xd}_{k,s}$ is an X-dimensional convolution with kernel size $k$ and stride $s$. We notate $s=2\downarrow$ for convolution with stride 2 to downsample and $s=2\uparrow$ for deconvolution with stride 2 to upsample the feature map or volume.

All in all, the fusion module for a 2D feature map and a 3D volume is formulated as
\begin{align}
\begin{split}
    \mathcal{F}\left(\mathbf{v}, \mathbf{f}\right) = 
    \Phi^{3d}_{3,1}\left(\Psi\left(\Gamma^{3d}\left(\mathbf{v}\right) + \Gamma^{2d}\left(\mathbf{f}\right)\right)\right)
\end{split}
\end{align}
where $+$ is the broadcasting sum and
\begin{align}
\begin{split}
    \Gamma^{Xd}\left(x\right) = {\footnotesize\text{Conv}}^{Xd}_{1,1}\left(\Psi\left(\gamma^{Xd}\left(x\right)\right)\right)
    \\
\end{split}\\
\begin{split}
    \Psi\left(x\right)=\footnotesize\text{LeakyReLU}_{0.1}\left(\footnotesize\text{BatchNorm}\left(x\right)\right).
\end{split}
\end{align}

The following subsection will describe the detailed usage of the fusion blocks in our model.

% \label{sec:methods}

% % % % % % % % % % % % % % % % % % % % % % % % % % % % % % % % 
\subsection{Image-coupled Volume Propagation}\label{sec:volume_propagation}
The 2D pipeline has a UNet structure, consisting of encoder layers to downsample the incoming feature maps, and decoder layers to recover intermediate resolutions and integrate the skip connections from the encoders as
\begin{align}
\begin{split}
    & \footnotesize\text{Enc}^{2d} \left(\mathbf{f}^\mathrm{Enc}_{i-1}\right) = \Phi^{2d}_{3, 1}\left(\Phi^{2d}_{3,1}\left(\Phi^{2d}_{3,2\downarrow}\left(\mathbf{f}^\mathrm{Enc}_{i-1}\right)\right)\right)
\end{split}\\
\begin{split}
     \footnotesize\text{Dec}^{2d} \left(\mathbf{f}^{\mathrm{Dec}}_{i+1}, \mathbf{f}^\mathrm{Enc}_{i}\right) 
    = \Phi^{2d}_{3,1}\left( \Phi^{2d}_{3,1}\left( \footnotesize\text{Skip}^{2d} \left(\mathbf{f}^{\mathrm{Dec}}_{i+1}, \mathbf{f}^\mathrm{Enc}_{i}\right)\right)\right)
\end{split}
\end{align}
where
\begin{align}
\begin{split}
    \footnotesize\text{Skip}^{Xd} \left(x^{\mathrm{Dec}}_{i+1}, x^\mathrm{Enc}_{i}\right) = \Phi^{Xd}_{1,1}\left(\Phi^{Xd}_{3,2\uparrow}\left(x^{\mathrm{Dec}}_{i+1}\right) \oplus x^\mathrm{Enc}_{i}\right)
\end{split}
\end{align}
and $\oplus$ denotes the concatenation along the feature axis.
By concatenating the skip connection from the encoder part, the decoder layers are able to keep the detailed features on the upper resolutions while handling semantic information coming from lower resolutions.

In order to evolve the cost volumes $v_i$, we keep a similar encoder and decoder structures with skip connections. In addition, the 3D UNet pipeline has incoming connections from the feature maps $f_i$.
The 2D-3D fusion is computed via broadcasting sum after the atrous blocks as
\begin{align}
\begin{split}
    &\footnotesize\text{Enc} ^{3d}\left(\mathbf{v}^\mathrm{Enc}_{i-1}, \mathbf{f}^\mathrm{Enc}_{i}\right) = \Phi^{3d}_{3,1}\left(\mathcal{F}\left(\Phi^{2d}_{3,2\downarrow}\left(\mathbf{v}^\mathrm{Enc}_{i-1}\right), \mathbf{f}^\mathrm{Enc}_{i}\right)\right)
\end{split} \\
\begin{split}
    &\footnotesize\text{Dec}^{3d}\left(\mathbf{v}^{\mathrm{Dec}}_{i+1}, \mathbf{v}^\mathrm{Enc}_{i}, \mathbf{f}^\mathrm{Dec}_{i}\right) = 
    \mathcal{F}\left(\footnotesize\text{Skip}^{3d} \left(\mathbf{v}^{\mathrm{Dec}}_{i+1}, \mathbf{v}^\mathrm{Enc}_{i} \right), \mathbf{f}^\mathrm{Dec}_{i}\right)
\end{split}
\end{align}

Since the last layers of the cost volumes have the largest resolution which dominantly exhausts GPU memory and computation power, we define a lighter $\footnotesize\text{Dec}^{3d}$ at the last layer without the atrous block as
\begin{align}
\begin{split}
    \mathbf{v}^\mathrm{Dec}_{0} =  {\footnotesize\text{Conv}}^{3d}_{3,1}\left(\Psi\left({\footnotesize\text{Conv}}^{3d}_{3,1}\left( \mathbf{v}^\mathrm{Dec}_{0, skip} \right)
     +{\footnotesize\text{Conv}}^{2d}_{3,1}\left(\mathbf{f}^\mathrm{Dec}_{0}\right)\right)\right)
\end{split}
\end{align}
where $\mathbf{v}^\mathrm{Dec}_{0, skip} = \Phi^{3d}_{3,2\uparrow}\left(\mathbf{v}^{\mathrm{Dec}}_{1}\right) \oplus \mathbf{v}^\mathrm{Enc}_{0}$.

\section{Experiment}
\label{sec:experiment}

\begin{table*}[t]\centering
  \begin{tabular}{ccc||ccc|cc}
\hline
& & & \multicolumn{3}{c|}{SceneFlow } & KITTI 15 &  KITTI 12 \\
\hline 
% Atrous & 2D path & Bad 1.0 & Bad 3.0 & EPE & Time(Sec) & GPU Mem(GB)\\
2D & GWC & Atrous & EPE & \makecell{time(s)} & \makecell{GPU Mem.\\(MB)} & EPE & EPE\\
\hline
\xmarknew & \xmarknew & \xmarknew &  0.639 & 0.139 & 3470 & 4.382 & 4.128 \\
\cmarknew & \xmarknew & \xmarknew &  $\mathbf{0.546}$ & 0.143 & 3594 & 4.384 & 3.942 \\
%\xmarknew & \xmarknew & \cmarknew &  0.601 & 0.161 & 4012 & 3.952 & 3.063 \\
\cmarknew & \xmarknew & \cmarknew &  $\mathbf{0.546}$ & 0.168 & 4162 & 4.607 & 3.932 \\
%\xmarknew & \cmarknew & \xmarknew & 0.578 & 0.155 & 4020 & 5.872 & 5.812 \\
\cmarknew & \cmarknew & \xmarknew & 0.569 & 0.159 & 4084 & 6.041 & 5.874 \\
%\xmarknew & \cmarknew & \cmarknew &  0.558 & 0.177 & 4504 & 3.929 & 4.293 \\
\cmarknew & \cmarknew & \cmarknew & 0.548 & 0.182 & 4596 & $\mathbf{3.755}$ & $\mathbf{3.599}$ \\
\hline 
\end{tabular}
  \caption{Ablation Study. GWC, Atrous, and 2D denote groupwise correlation, atrous convolution, and 2D propagation.}
  \label{tab:ablation}
\end{table*}

% % % % % % % % % % % % % % % % % % % % % % % % % % % % % % % % 
We evaluate our model on the public datasets: Sceneflow\cite{mayer2016large}, KITTI 2012\cite{geiger2012we}, ETH3D\cite{schops2017multi}, and Middlebury V3\cite{scharstein2014high}.
Sceneflow includes 35,454 synthetic image pairs for training and respectively 4,370 pairs for testing. The training includes the FlyingThings, Monkaa, and Driving datasets, each consisting of 22390, 8664, and 4400 image pairs. 
The KITTI 2012 dataset is comparably small but includes real world driving scenarios. It consists of 194 image pairs for training and 194 image pairs for testing. Ground truth data was measured by a $360^{\circ}$ laser scanner and is sparser than the original image data. The ETH3D dataset contains only grayscale images, covering indoor and outdoor scenes, and consist of 27 and 20 image pairs for training and testing. Sparse ground truth results were obtained by laser scans.
The Middlebury V3 dataset contains 15 indoor scenes for each training and testing and each scene includes 3 different image resolutions with relatively dense ground truths. A few image pairs are captured with asymmetric lighting conditions to increase the matching challenge.

\paragraph*{Evaluation Metrics:} We use the End-Point-Error (EPE), representing the averaged absolute difference between the computed result and ground truth, and the percentage of outliers, denoted as $n$-noc, $n$-all, or Bad $n$ where a pixel is discarded if the absolute value of the error is over the threshold $n$. On the official KITTI 2012 benchmark, $n$-noc denotes the percentage of outliers only for non-occluded pixels and $n$-all is for all pixels.

\subsection{Training Details}\label{sec:training}
Our implementation relies on PyTorch and was evaluated on an NVIDIA RTX 3090 GPU.
For the ablation study (Table \ref{tab:ablation}) and the evaluation on the Sceneflow test set (Table \ref{tab:sceneflow}), we trained our model only with the FlyingThings dataset, cropped to the image size of $576 \times540$ without any image augmentation during training.
The model is trained for 60 epochs with a batch size of 2 using the Adam optimizer, beginning with a learning rate of $1e^{-3}$ and dropping by half at the 25th, 30th, 35th, 38th, 41th, 44th, and 47th epochs, and gradient clipping\cite{zhang2019gradient} with the maximum norm of 0.1.

For the evaluation on other benchmarks (Table \ref{tab:kitti12} and \ref{tab:eth3d}), the model is pre-trained on the entire Sceneflow training set, with images cropped to the size of $576\times384$ and with data augmentation.
Chromatic augmentation is applied with up to 30$\%$ on brightness, contrast, and saturation for both the left and right images in the symmetric and asymmetric ways with a ratio of 4 to 1.
To enhance image completion performance, we adopt occlusion augmentation on the right image before cropping when building training batches\cite{tankovich2021hitnet}.
The size of replaced cropped patch ranges randomly between $50\times50$ and $100\times100$.
All other settings are identical to the training for the SceneFlow test set and our ablation study, except for the batch size, which is 4 instead of 2.

For the KITTI 2012 benchmark, the pre-trained model is finetuned on the KITTI 2012 training set with the previously described occlusion augmentation.
To prevent the model from overfitting to the limited number of training data, the images are resized by a scaling factor between 1 and 2 and cropped to fit the size of $1241\times376$, resembling to the full resolution of the test images.
The model is retrained for 240 epochs with a batch size of 2, a maximum norm of 0.1 for gradient clipping and the Adam optimizer. Beginning with a learning rate of $1e^{-3}$ it is divided by two at epoch 50, 100, 150, and 200.

For the evaluation on the ETH3D benchmark, we use the training sets of ETH3D, Middlebury V3, ORStereoDataset\cite{hu2021orstereo}, Sintel\cite{butler2012naturalistic}, and InStereo2K\cite{bao2020instereo2k} with cropped images to a resolution of $576\times540$ pixels.
We downscale the ORStereoDataset to half of the original resolution and remove samples if the maximum disparity is over 800. To balance the ratio of the datasets during each epoch, a single frame is picked from each sequence of the Sintel dataset, and only one hundred randomly selected samples are chosen from InStereo2K.
All training sets are enriched by asymmetric chromatic augmentation and occlusion augmentation.
To handle robustly lens flare effects within the ETH3D test sets, resulting from  filming against the sun, image brightness was adjusted to 70$\%$ -- 200$\%$ within the chromatic augmentation.
The model is finetuned for 240 epochs with a batch size of 2, a maximum norm for gradient clipping of 0.01, and a learning rate starting from $1e^{-3}$ and dropping by two at the 100th and 200th epochs.

 \begin{table}[ht]
\begin{center}
{
% \resizebox{\columnwidth}{!}
{%
\begin{tabular}{cl||ccc}
\hline
& & \multicolumn{3}{c}{SecneFlow} \\
\hline
Input & Model & EPE & \makecell{time(s)} & \makecell{GPU Mem.\\(MB)}\\ [3pt]
\hline 
\multirow{3}{*}{\rotatebox[origin=r]{90}{\makecell{\small{Cropped}\\ \small{(940$\times$512)}}}} 
& GWCNet\cite{guo2019group} & 0.77 & $\mathbf{0.16}$ & $\underline{5208}$\\[3pt]
& ACVNet\cite{xu2022attention} & $\mathbf{0.48}$ & $0.20$ & 5940\\[3pt]
& ICVP & $ - $ & $\underline{0.17}$ & $\mathbf{4426}$\\[3pt]
\hline 
\multirow{3}{*}{\rotatebox[origin=r]{90}{\makecell{ \small{Full} \\ \small{(940×540)}}}}
& PSMNET\cite{chang2018pyramid} & 1.09 & 0.32 & $\mathbf{4358}$\\[3pt]
& LEA\cite{cheng2020hierarchical} & 0.78 & 0.50 & 6658\\[3pt]
& ICVP & $\underline{0.55}$ & $\mathbf{0.18}$ & $\underline{4596}$\\[3pt]
\hline 
\end{tabular}
}
}
\end{center}
\caption{
Quantitative results on SceneFlow finalpass for methods using 3D convolutions. Some models do not support the full resolution of SceneFlow data. For a fair comparison of runtime and GPU memory consumption we split the table.
}
\label{tab:sceneflow}
\end{table}

% 0.17
% 4436MiB
% \begin{table}[t]
% \begin{center}
% {%
% \begin{tabular}{l||ccccccc}
% \hline
% & \multicolumn{7}{c}{KITTI 2012} \\
% \hline 
%  & 2-noc & 2-all & 3-noc & 3-all & 4-noc & 4-all & time(s)\\
% \hline
% OptStereo & 1.91 & 2.51 & 1.2 & 1.61 & 0.92 & 1.24 & 0.1 \\
% NLCA-Net v2 & 1.83 & 2.34 & 1.11 & 1.46 & 0.83 & 1.09 & 0.67 \\
% LaC+GwcNet & 1.89 & 2.43 & 1.13 & 1.49 & 0.84 & 1.1 & 0.65 \\
% LEAStereo & 1.9 & 2.39 & 1.13 & 1.45 & 0.83 & 1.08 & 0.3 \\
% GANet-deep & 1.89 & 2.5 & 1.19 & 1.6 & 0.91 & 1.23 & 1.8 \\
% CREStereo & 1.72 & 2.18 & 1.14 & 1.46 & 0.9 & 1.14 & 0.4 \\
% ACVNet & 1.83 & 2.34 & 1.13 & 1.47 & 0.86 & 1.12 & 0.2 \\
% LaC+GANet & 1.72 & 2.26 & 1.05 & 1.42 & 0.8 & 1.09 & 1.8 \\
% PCWNet & 1.69 & 2.18 & 1.04 & 1.37 & 0.78 & 1.01 & 0.44 \\
% \hline
% ours & 1.72 & 2.21 & 1.06 & 1.39 & 0.8 & 1.05 & 0.17 \\
% \hline 
% \end{tabular}
% }
% \end{center}
% \caption{
% aaa
% }
% \label{tab:comparison}
% \end{table}

\begin{table}[ht]
\begin{center}
{
\resizebox{\columnwidth}{!}
{%
\begin{tabular}{l||ccccc}
\hline
& \multicolumn{5}{c}{KITTI 2012} \\
\hline 
Method & 3-noc & 3-all & 4-noc & 4-all & time(s)\\
\hline
GANet-deep\cite{zhang2019ga} & 1.19 & 1.6 & 0.91 & 1.23 & 1.8 \\
ACVNet\cite{xu2022attention} & 1.13 & 1.47 & 0.86 & 1.12 & $\underline{0.2}$ \\
CREStereo\cite{li2022practical} & 1.14 & 1.46 & 0.9 & 1.14 & 0.4 \\
NLCA-Net v2\cite{rao2022rethinking} & 1.11 & 1.46 & 0.83 & 1.09 & 0.67 \\
LEAStereo\cite{cheng2020hierarchical} & 1.13 & 1.45 & 0.83 & 1.08 & 0.3 \\
LaC+GwcNet\cite{liu2022local} & 1.13 & 1.49 & 0.84 & 1.1 & 0.65 \\
LaC+GANet\cite{liu2022local} & $\underline{1.05}$ & 1.42 & \underline{0.8} & 1.09 & 1.8 \\
PCWNet\cite{shenpcw2022pcw} & $\mathbf{1.04}$ & $\mathbf{1.37}$ & $\mathbf{0.78}$ & $\mathbf{1.01}$ & 0.44 \\
\hline
Ours & 1.06 & $\underline{1.39}$ & $\underline{0.8}$ & $\underline{1.05}$ & $\mathbf{0.17}$ \\
\hline 
\end{tabular}
}
}
\end{center}
\caption{
Quantitative results on KITTI 2012 dataset
}
\label{tab:kitti12}
\end{table}

\begin{table}[ht]
\begin{center}
{%
    % \resizebox{\columnwidth}{!}
    {%

        \begin{tabular}{l||ccc}
        \hline
        & \multicolumn{3}{c}{ETH3D} \\
        \hline 
        Method & Bad 2.0 & Bad 4.0 & time(s)\\
        \hline
        StereoDRNet\cite{chabra2019stereodrnet} & 1.66 & 0.5 & $\underline{0.15}$ \\
        HITNet\cite{tankovich2021hitnet} & 1.01 & 0.31 & $\mathbf{0.01}$ \\
        AdaStereo\cite{song2021adastereo} & 0.76 & 0.25 & 0.4 \\
        ACVNet\cite{xu2022attention} & 0.75 & 0.3 & 0.2 \\
        DIP-Stereo\cite{zheng2022dip} & 0.68 & 0.28 & - \\
        RAFT-Stereo\cite{lipson2021raft} & 0.56 & 0.22 & 0.81 \\
        CREStereo\cite{li2022practical} & $\mathbf{0.29}$ & $\mathbf{0.12}$ & $0.76$ \\
        %CREStereo\cite{li2022practical} & $\mathbf{0.29}$ & $\mathbf{0.12}$ & $0.76^{*}$ \\
        \hline
        Ours & $\underline{0.53}$ & $\underline{0.16}$ & $\underline{0.15}$ \\
        \hline 
        \end{tabular}
        }
    }
\end{center}
\caption{
Quantitative results on ETH3D benchmark.
%* denotes the runtime tested on our machine (RTX 3090).
}
\label{tab:eth3d}
\end{table}

\begin{figure*}[t]
\centering
\begin{subfigure}[t]{0.225\textwidth}
    \includegraphics[width=\textwidth]  
    {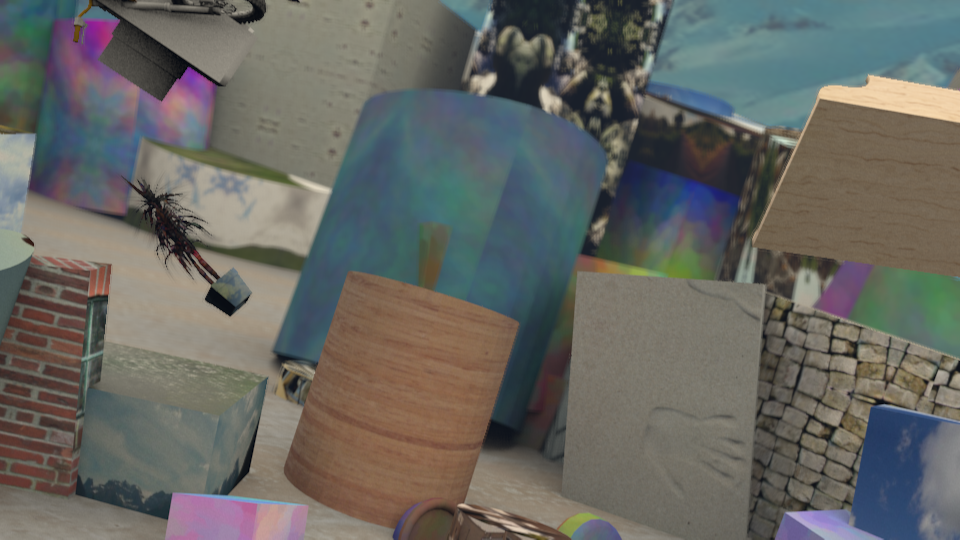}
    \includegraphics[width=\textwidth]
    {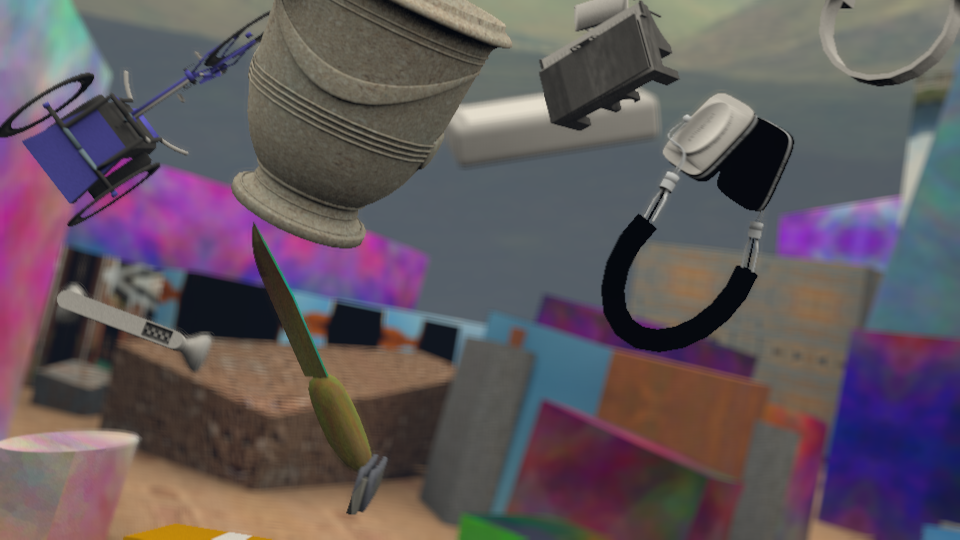}
    \includegraphics[width=\textwidth]
    {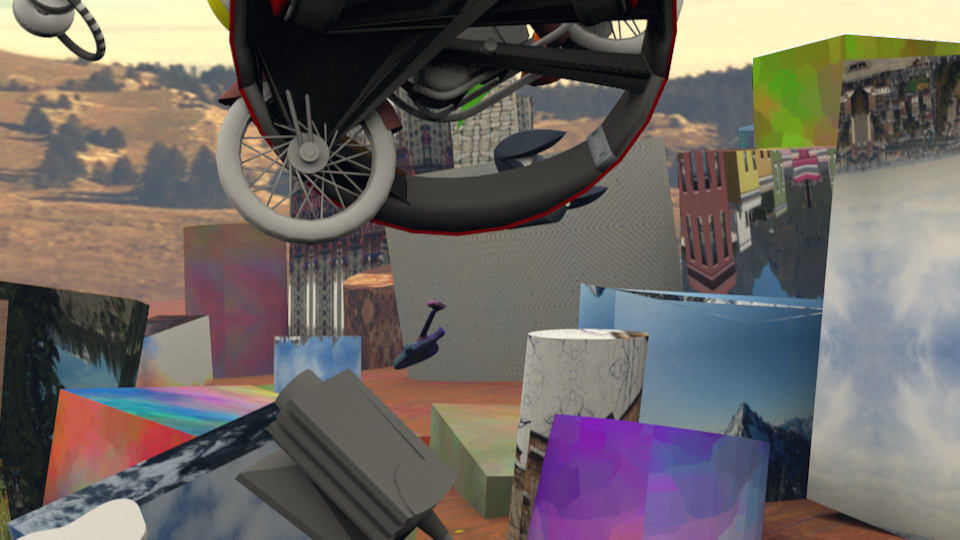}
    \caption{Left Image}
\end{subfigure}
\begin{subfigure}[t]{0.225\textwidth}
    \includegraphics[width=\textwidth]  
    {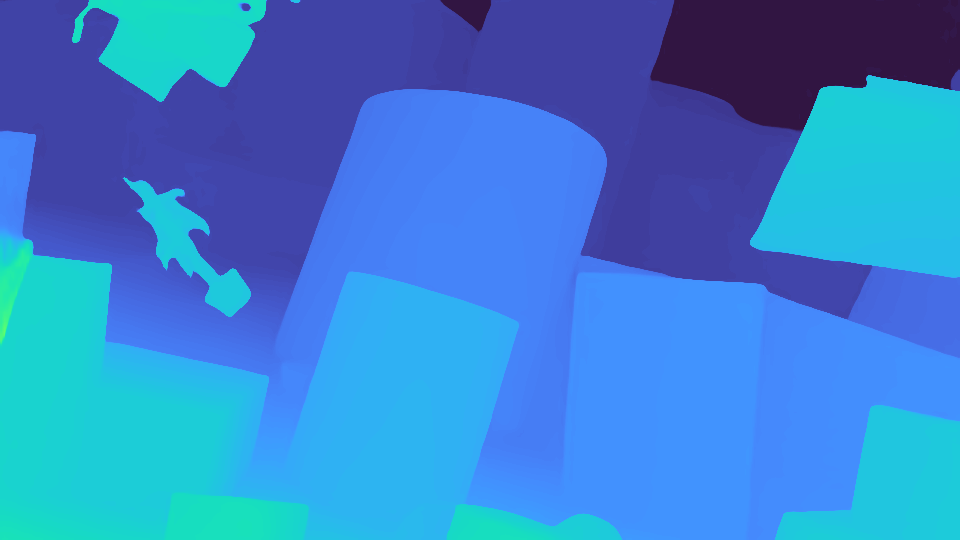}
    \includegraphics[width=\textwidth]
    {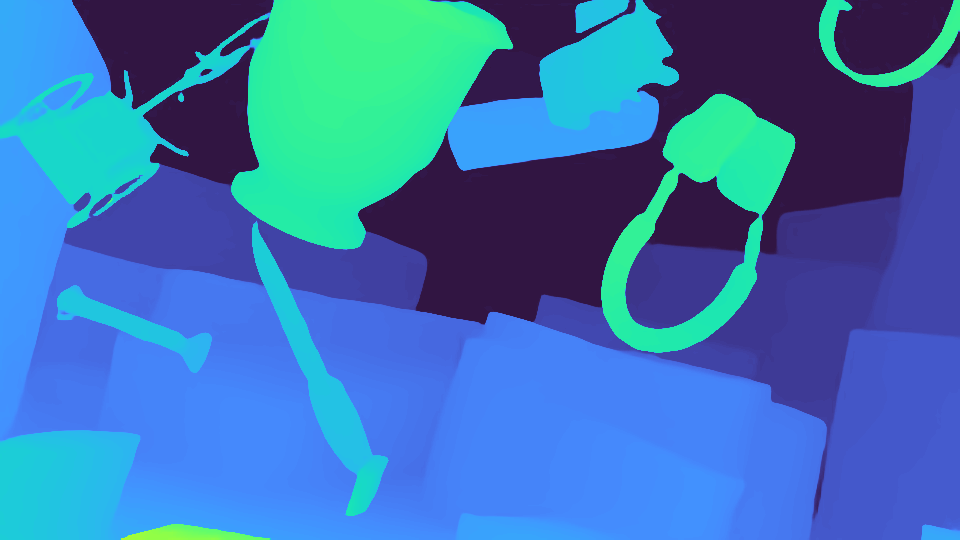}
    \includegraphics[width=\textwidth]
    {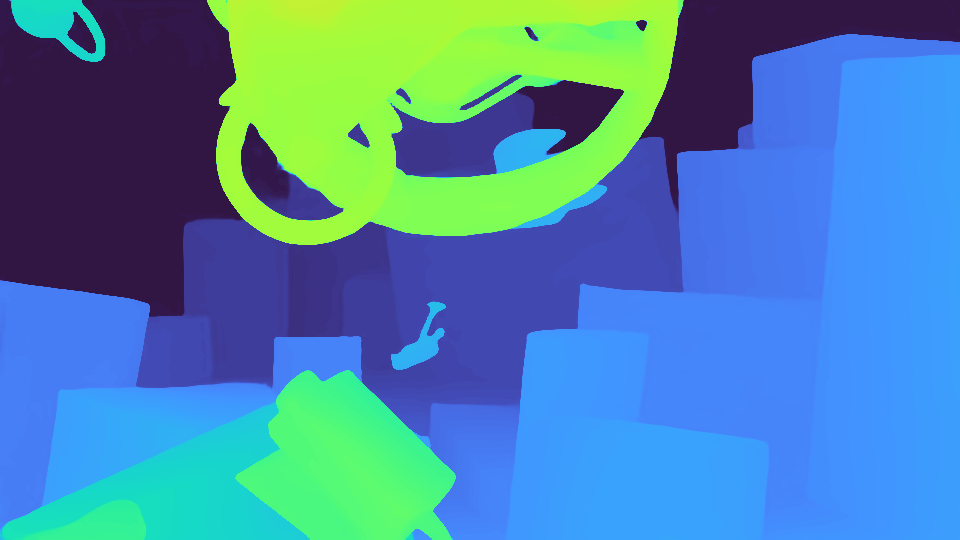}
    \caption{LEAStereo\cite{cheng2020hierarchical}}
\end{subfigure}
\begin{subfigure}[t]{0.225\textwidth}
    \includegraphics[width=\textwidth]  
    {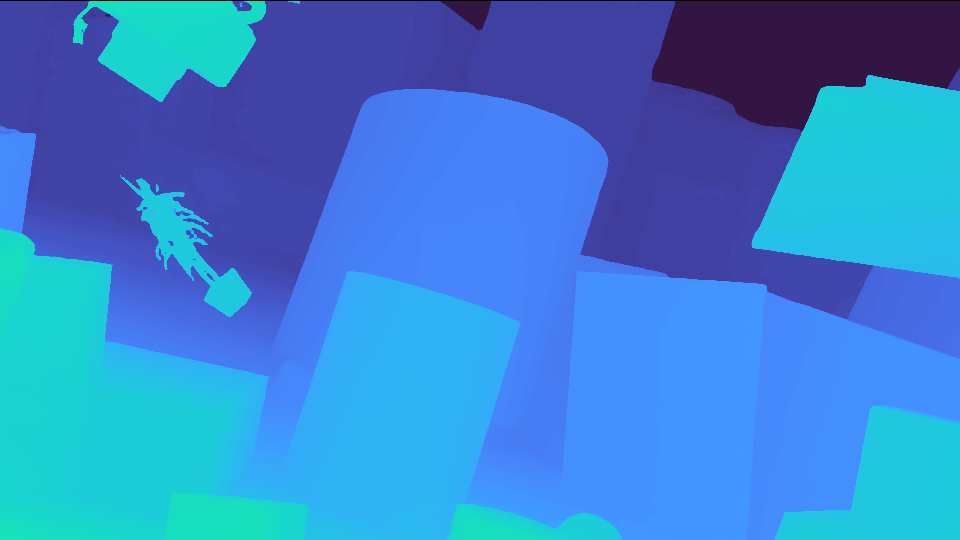}
    \includegraphics[width=\textwidth]
    {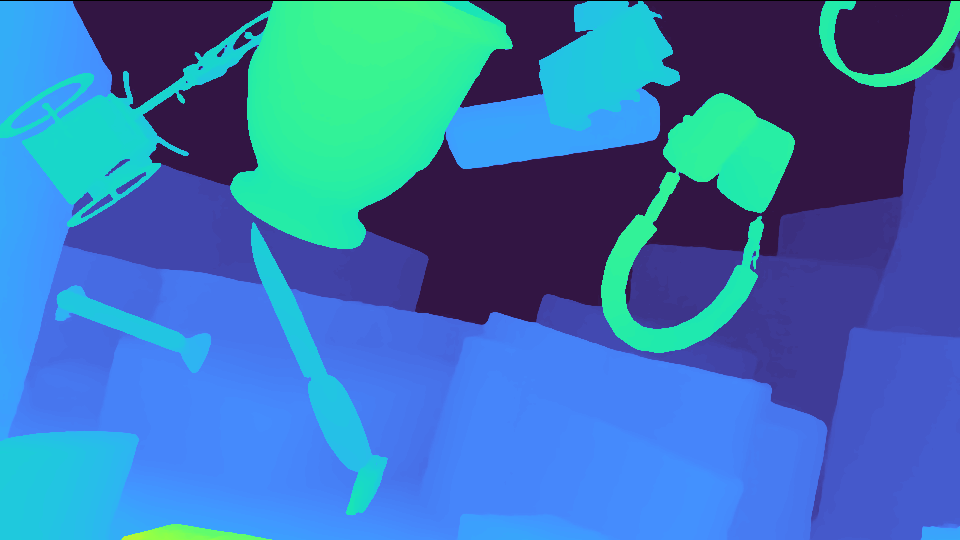}
    \includegraphics[width=\textwidth]
    {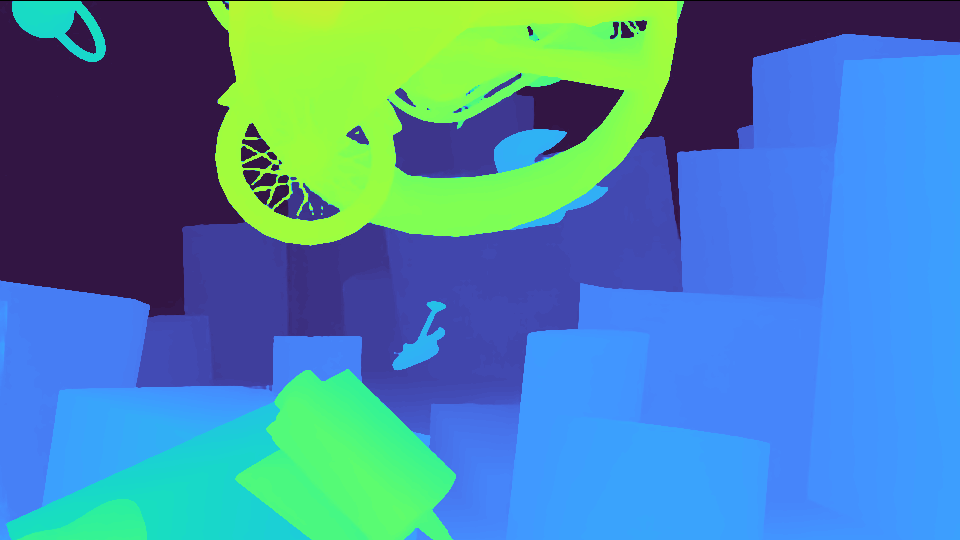}
    \caption{Ours}
\end{subfigure}
\begin{subfigure}[t]{0.225\textwidth}
    \includegraphics[width=\textwidth]  
    {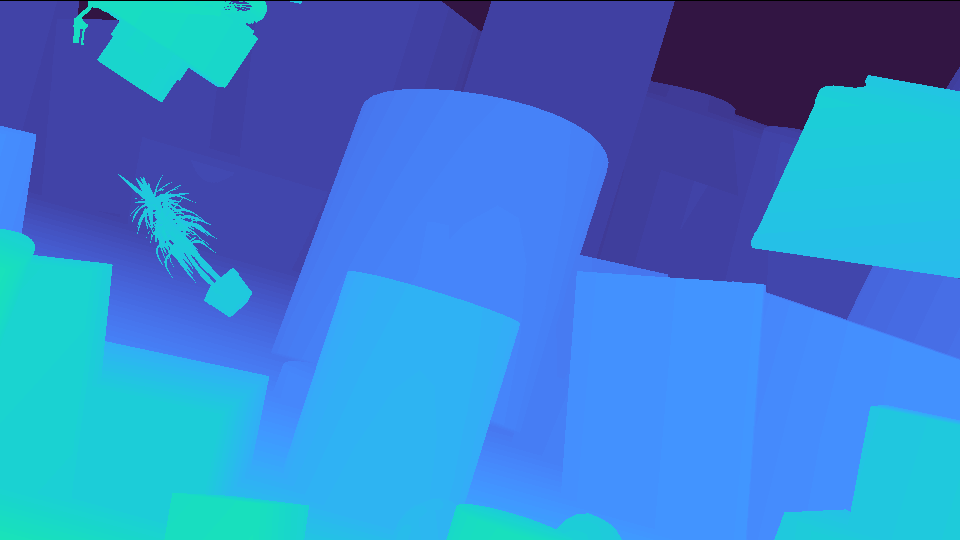}
    \includegraphics[width=\textwidth]
    {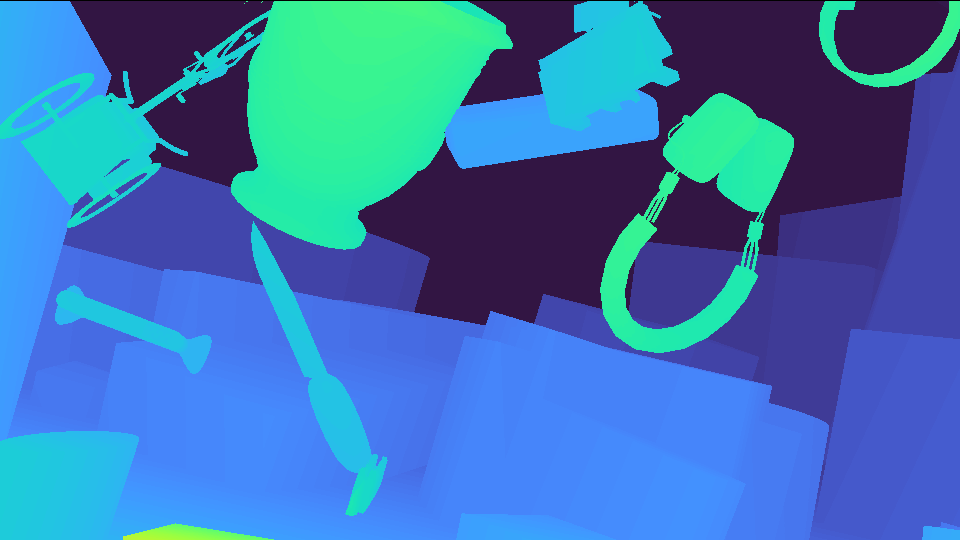}
    \includegraphics[width=\textwidth]
    {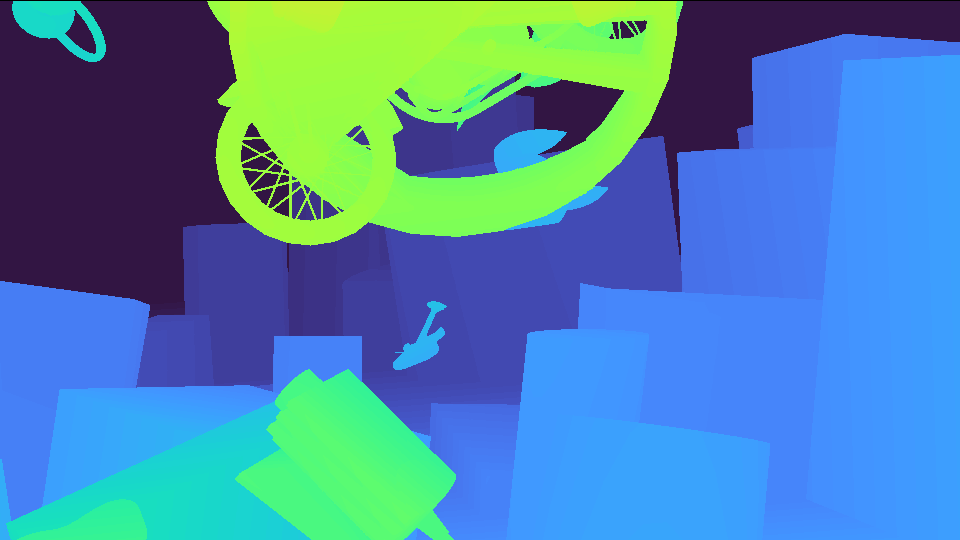}
    \caption{Ground Truth}
\end{subfigure}
\caption{Qualitative comparison on Sceneflow finalpass}
\label{fig:sceneflow}
\end{figure*}

\begin{figure*}[t]
\centering
\begin{subfigure}[t]{\dimexpr0.23\textwidth+15pt\relax}
    \makebox[15pt]{\raisebox{15pt}{\rotatebox[origin=c]{90}{\tiny{Image}}}}%
    \includegraphics[width=\dimexpr\linewidth-15pt\relax]
    {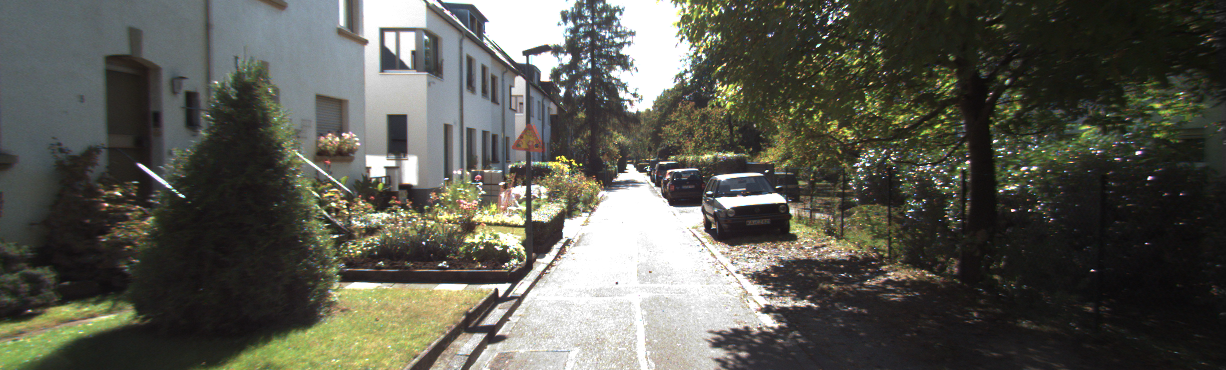}
    \makebox[15pt]{\raisebox{15pt}{\rotatebox[origin=c]{90}{\tiny{LEAStereo\cite{cheng2020hierarchical}}}}}%
    \includegraphics[width=\dimexpr\linewidth-15pt\relax]
    {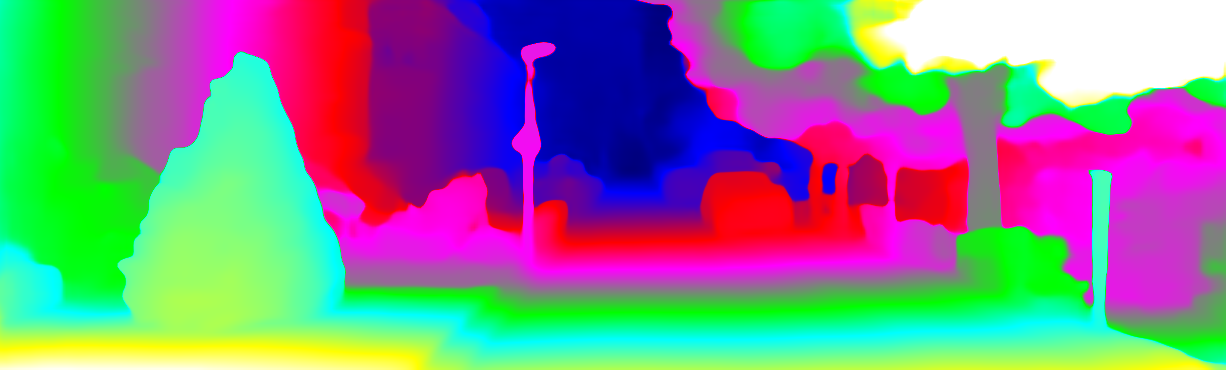}
    \makebox[15pt]{\raisebox{15pt}{\rotatebox[origin=c]{90}{\tiny{ACVNet\cite{xu2022attention}}}}}%
    \includegraphics[width=\dimexpr\linewidth-15pt\relax]
    {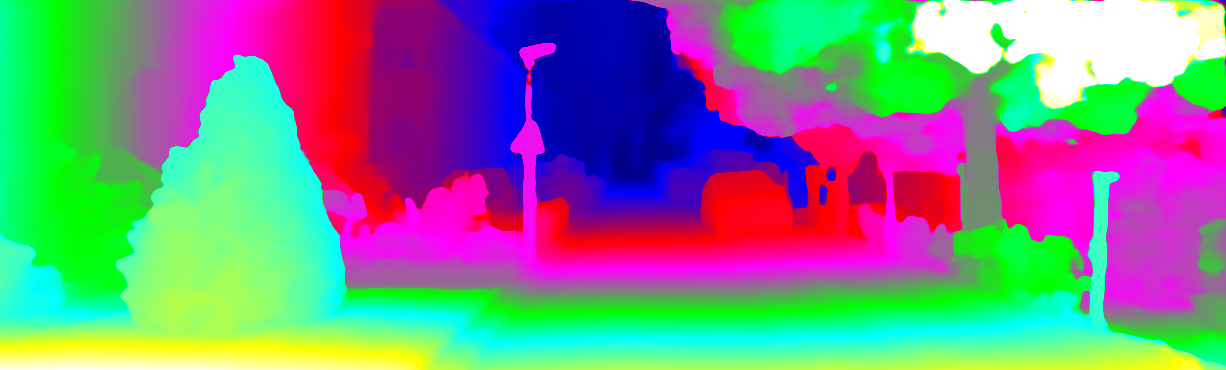}
    \makebox[15pt]{\raisebox{15pt}{\rotatebox[origin=c]{90}{\tiny{LaC+GANet\cite{liu2022local}}}}}%
    \includegraphics[width=\dimexpr\linewidth-15pt\relax]
    {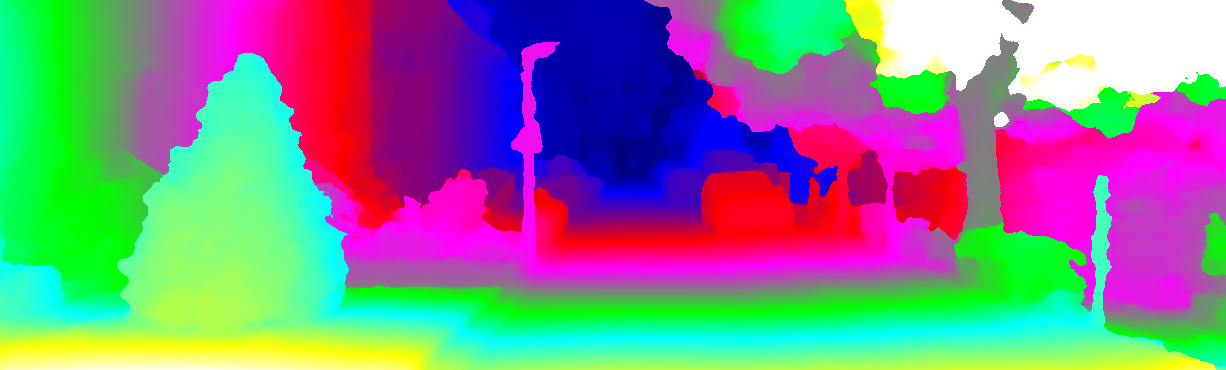}
    \makebox[15pt]{\raisebox{15pt}{\rotatebox[origin=c]{90}{\tiny{PCWNet\cite{shenpcw2022pcw}}}}}%
    \includegraphics[width=\dimexpr\linewidth-15pt\relax]
    {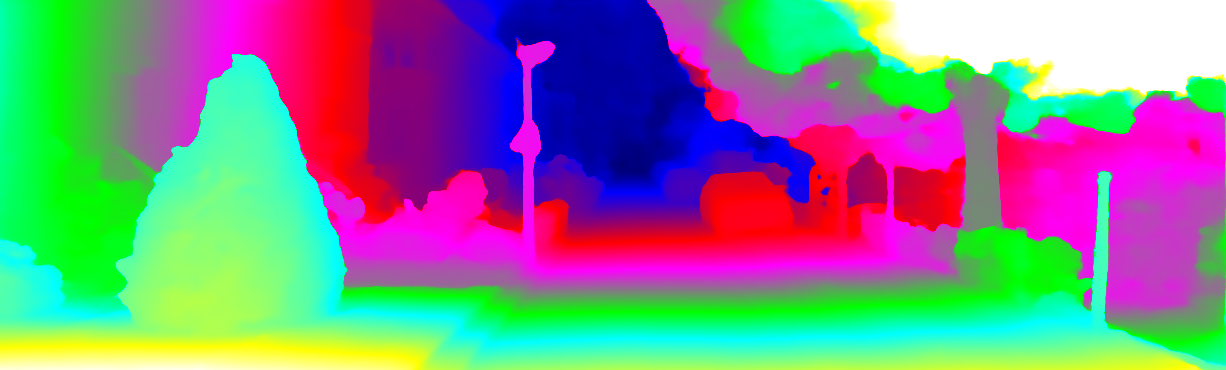}
    \makebox[15pt]{\raisebox{15pt}{\rotatebox[origin=c]{90}{\tiny{Ours}}}}%
    \includegraphics[width=\dimexpr\linewidth-15pt\relax]
    {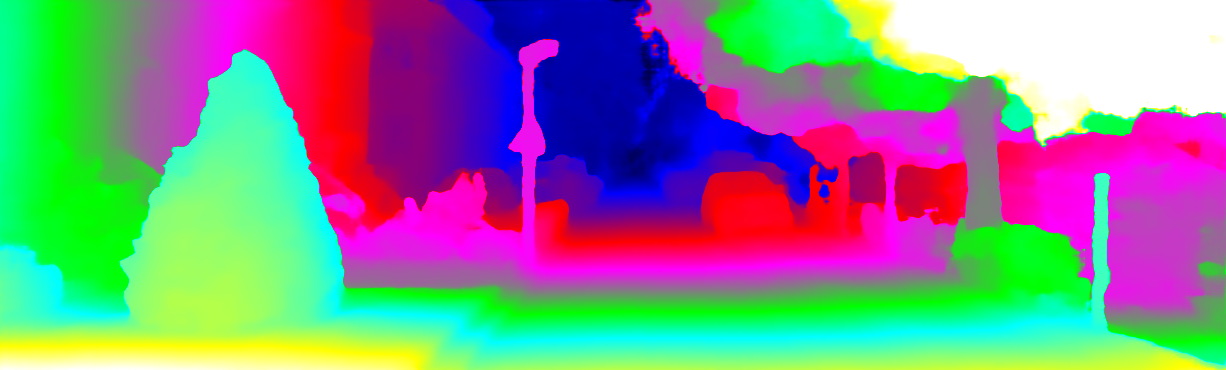}
\end{subfigure}
\begin{subfigure}[t]{0.23\textwidth}
    \includegraphics[width=\textwidth]  
    {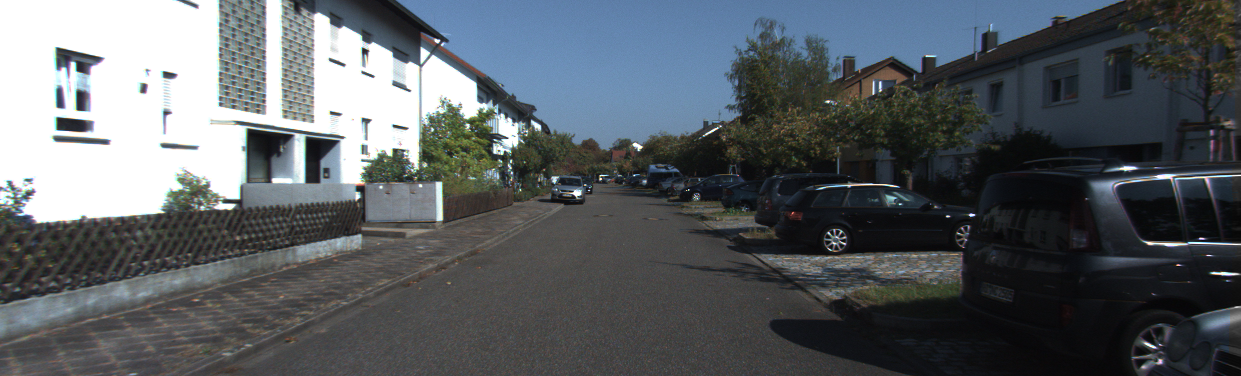}
    \includegraphics[width=\textwidth]
    {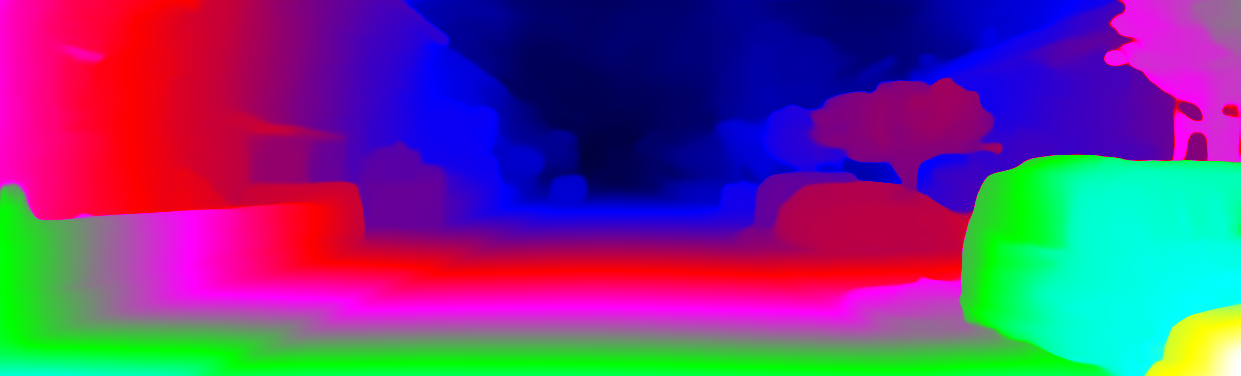}
    \includegraphics[width=\textwidth]
    {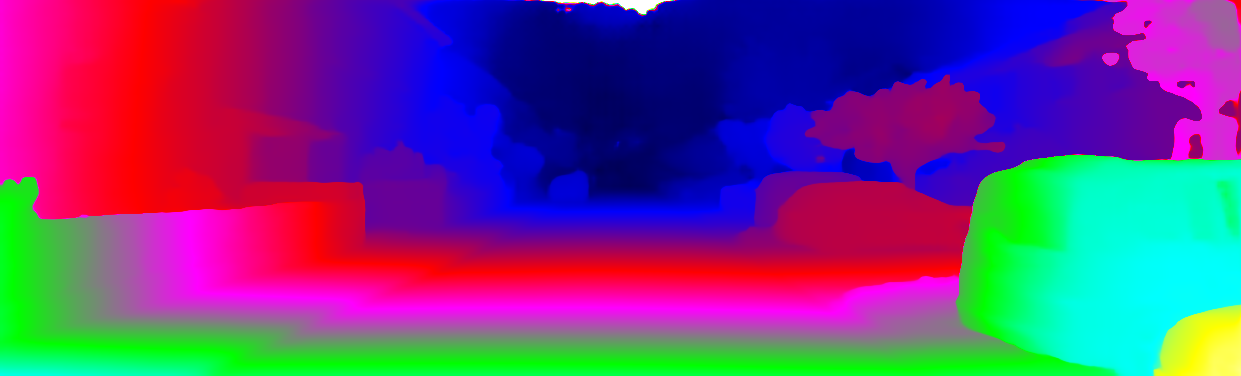}
    \includegraphics[width=\textwidth]
    {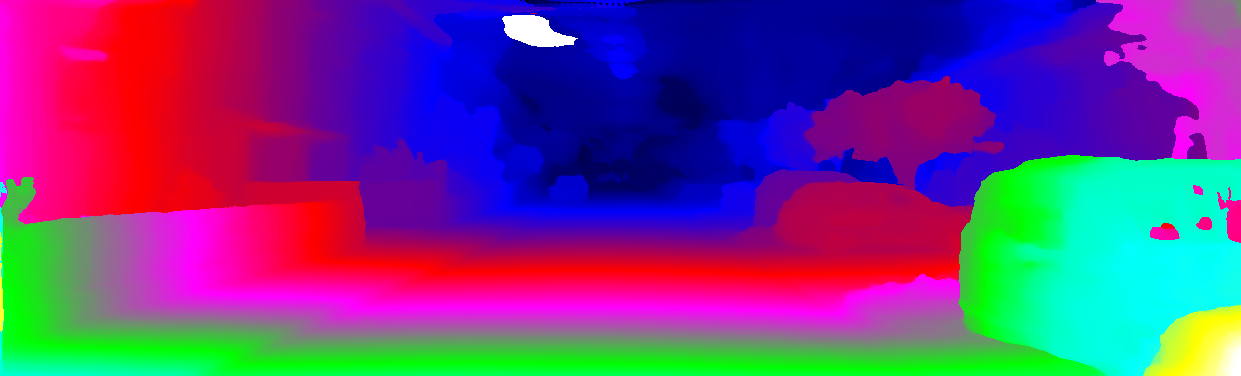}
    \includegraphics[width=\textwidth]
    {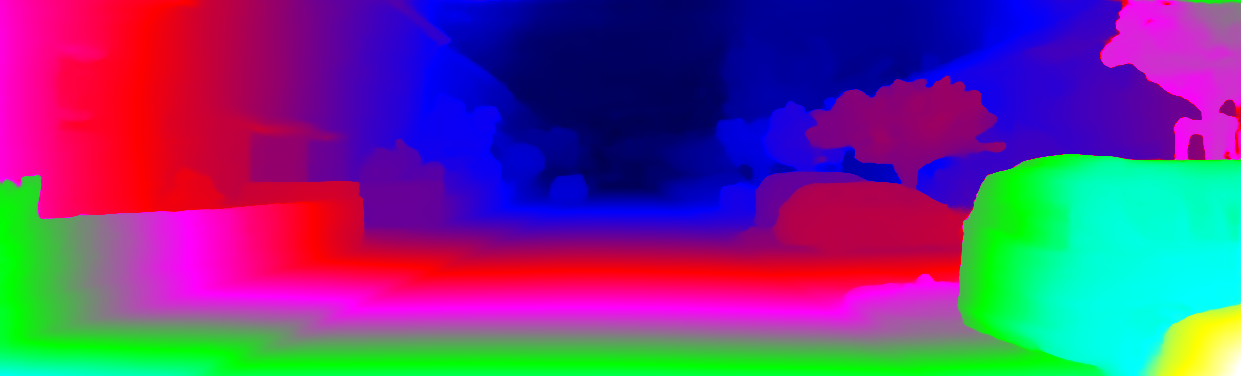}
    \includegraphics[width=\textwidth]
    {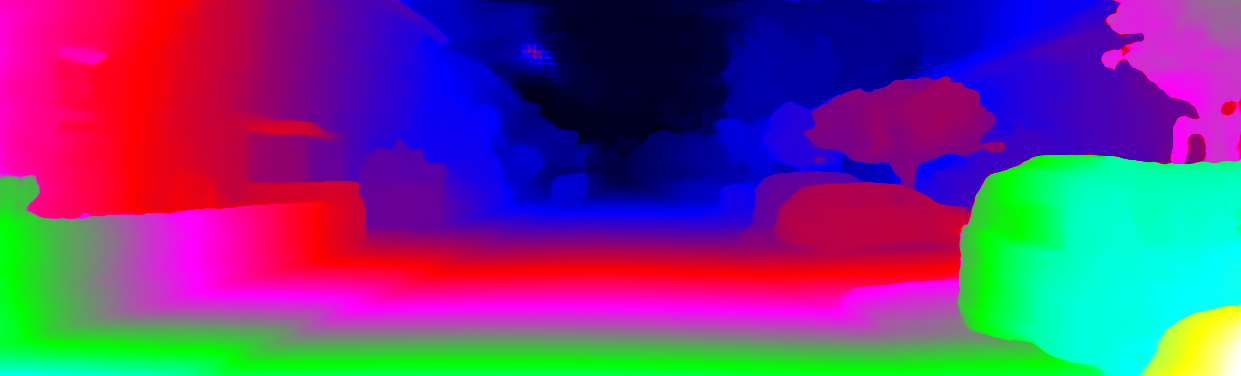}
\end{subfigure}
\begin{subfigure}[t]{0.23\textwidth}
    \includegraphics[width=\textwidth]  
    {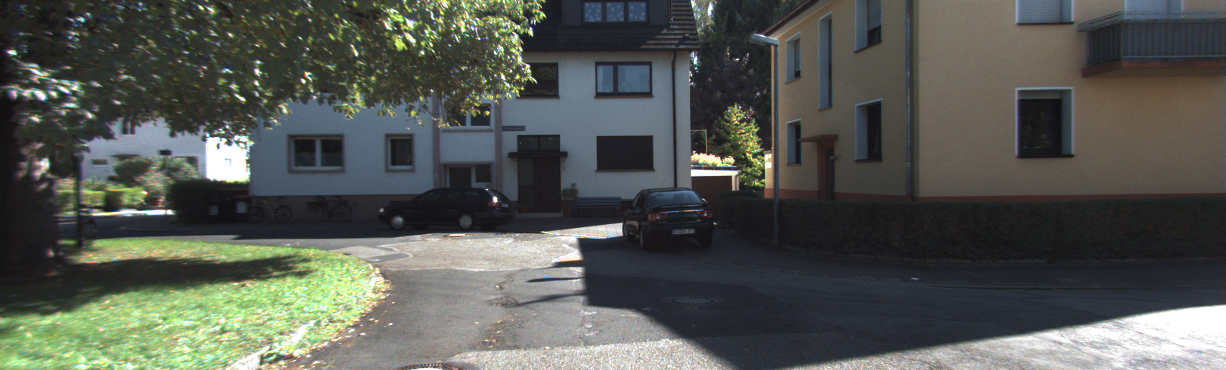}
    \includegraphics[width=\textwidth]
    {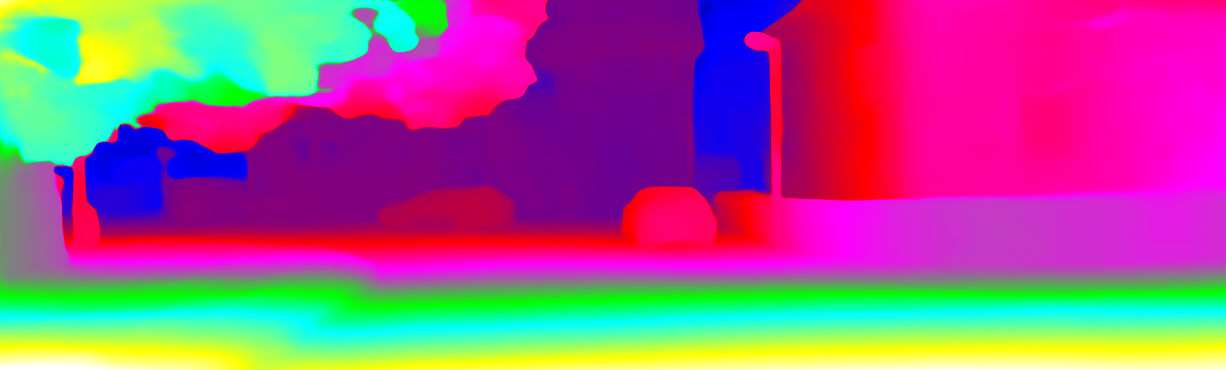}
    \includegraphics[width=\textwidth]
    {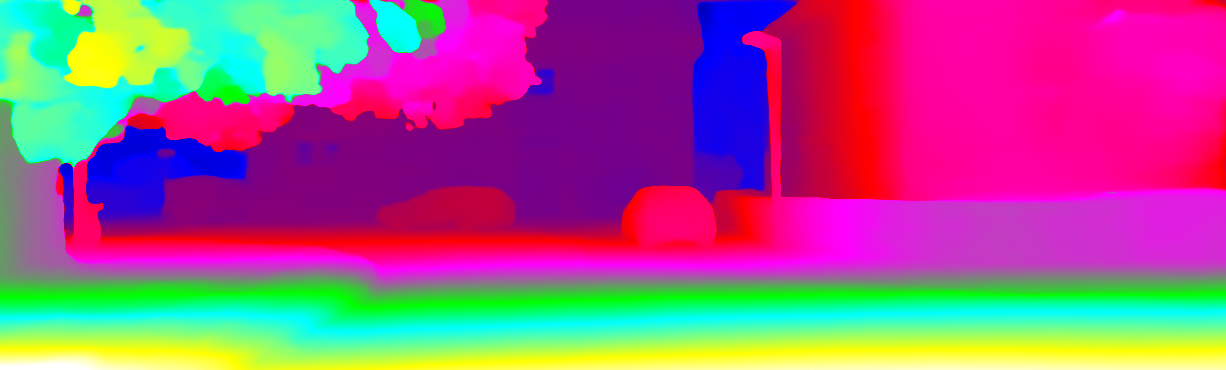}
    \includegraphics[width=\textwidth]
    {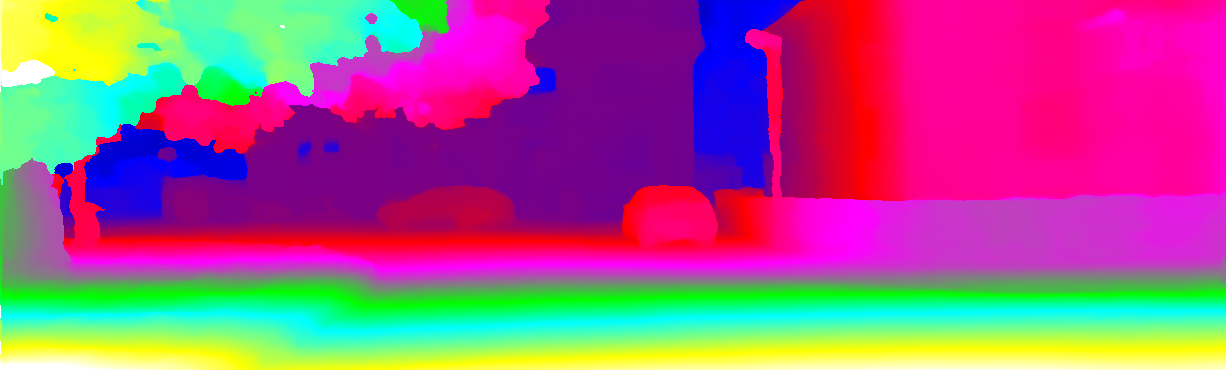}
    \includegraphics[width=\textwidth]
    {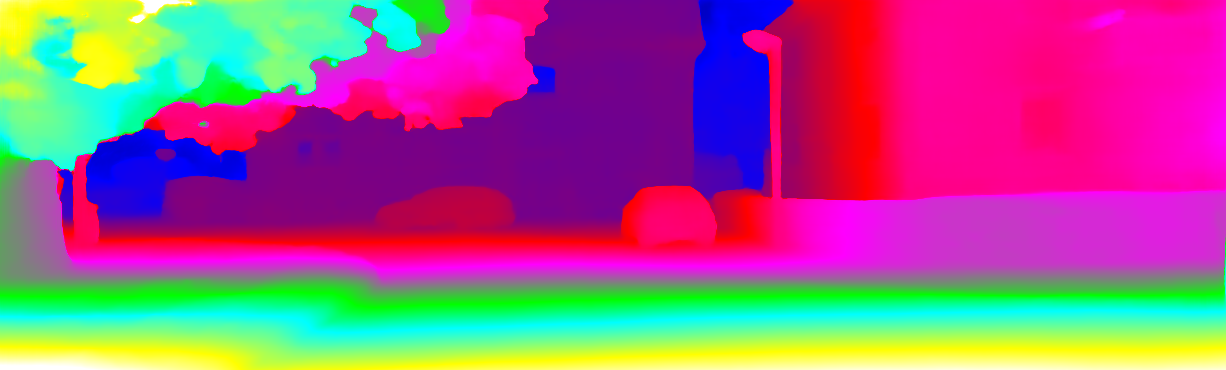}
    \includegraphics[width=\textwidth]
    {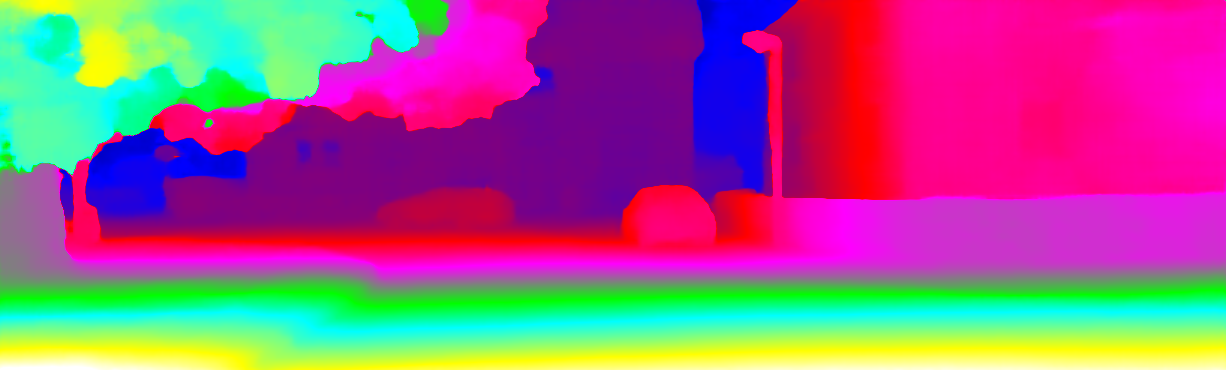}
\end{subfigure}
\begin{subfigure}[t]{0.23\textwidth}
    \includegraphics[width=\textwidth]  
    {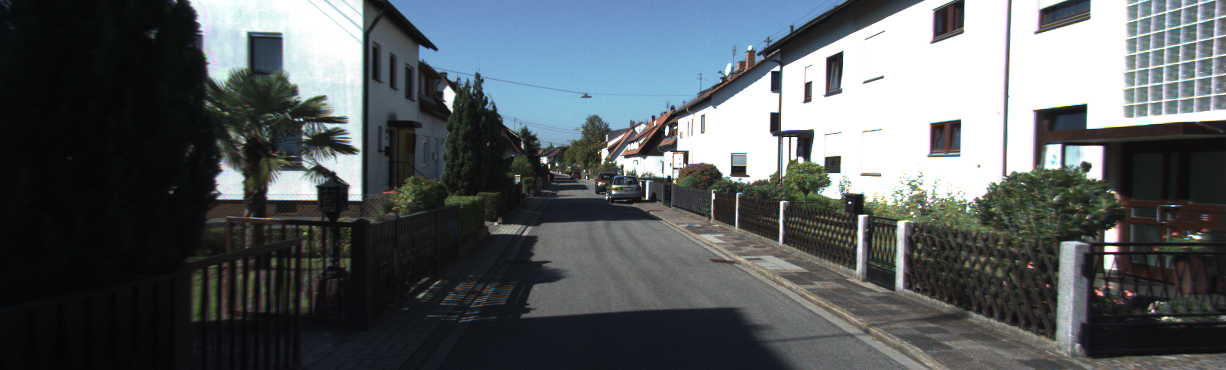}
    \includegraphics[width=\textwidth]
    {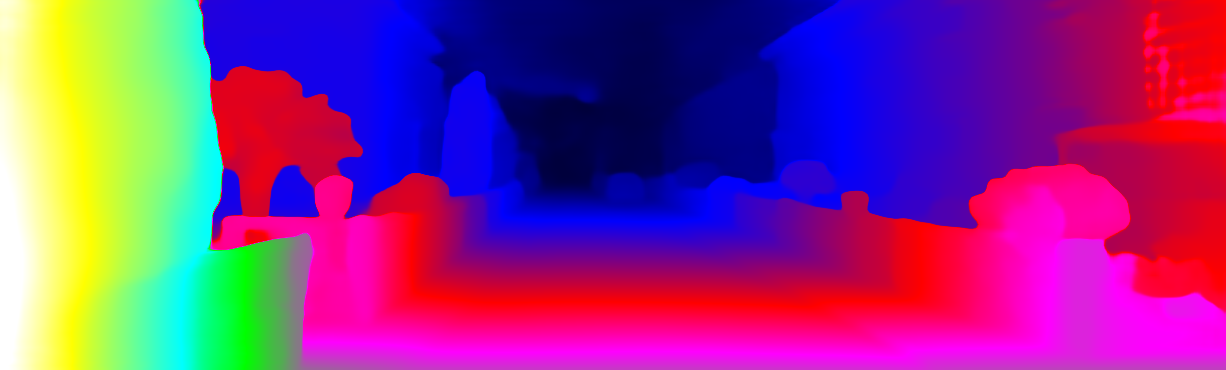}
    \includegraphics[width=\textwidth]
    {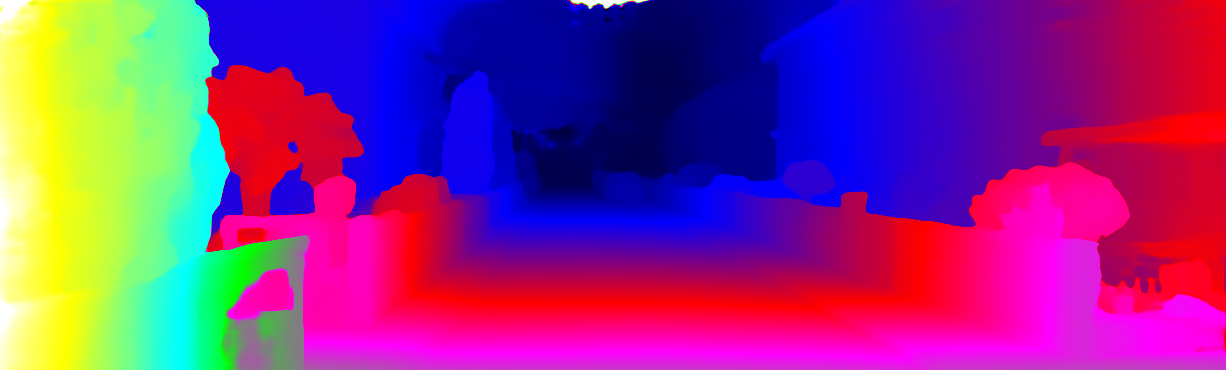}
    \includegraphics[width=\textwidth]
    {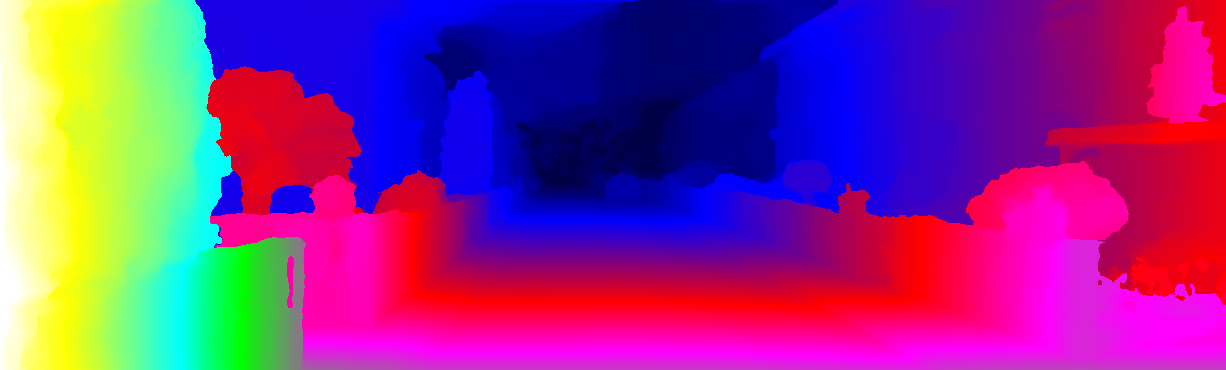}
    \includegraphics[width=\textwidth]
    {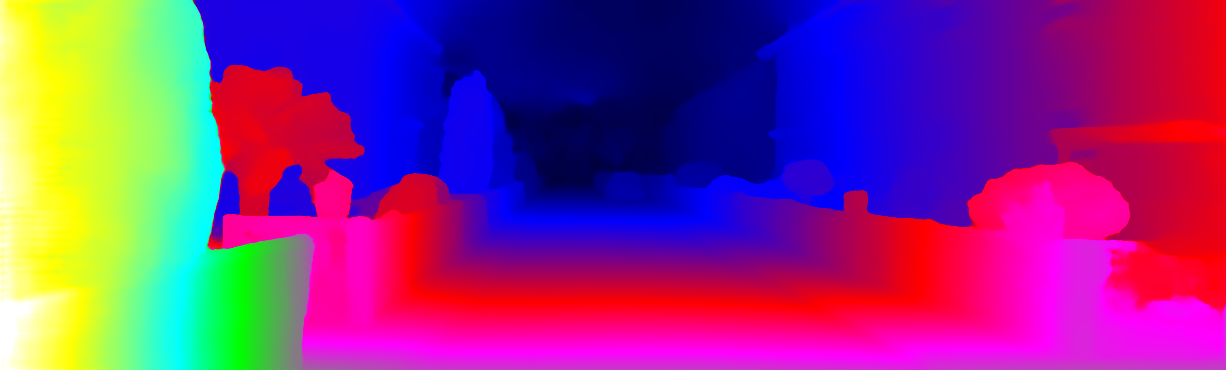}
    \includegraphics[width=\textwidth]
    {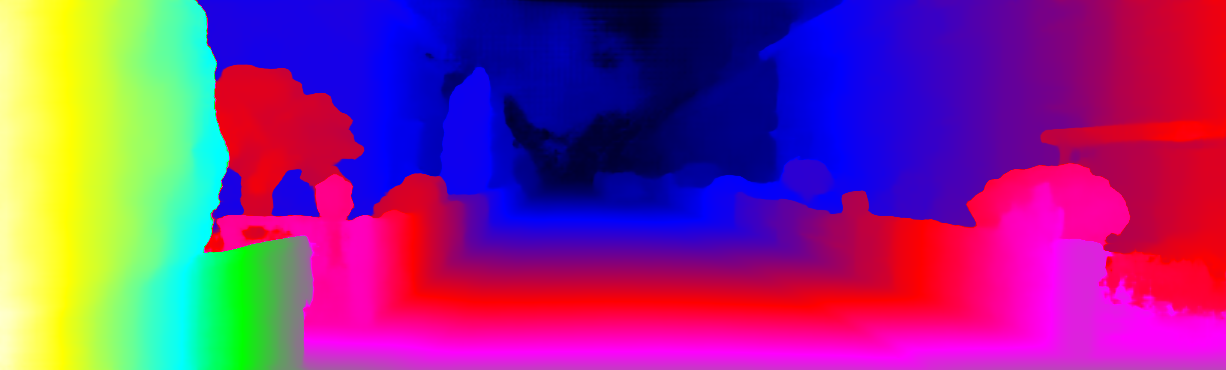}
\end{subfigure}
% \begin{subfigure}[t]{0.315\textwidth}
%     \includegraphics[width=\textwidth]  
%     {imgs/000008_10_img.png}
%     \includegraphics[width=\textwidth]
%     {imgs/000008_10_lea.png}
%     \includegraphics[width=\textwidth]
%     {imgs/000008_10_ours.png}
%     \includegraphics[width=\textwidth]
%     {imgs/000008_10_err.png}
% \end{subfigure}
\caption{Qualitative comparison on KITTI 2012}
\label{fig:kitti2012}
\end{figure*}

\begin{figure*}[ht]
\begin{subfigure}{0.19\linewidth}
  \centering
  \includegraphics[width=.99\linewidth]{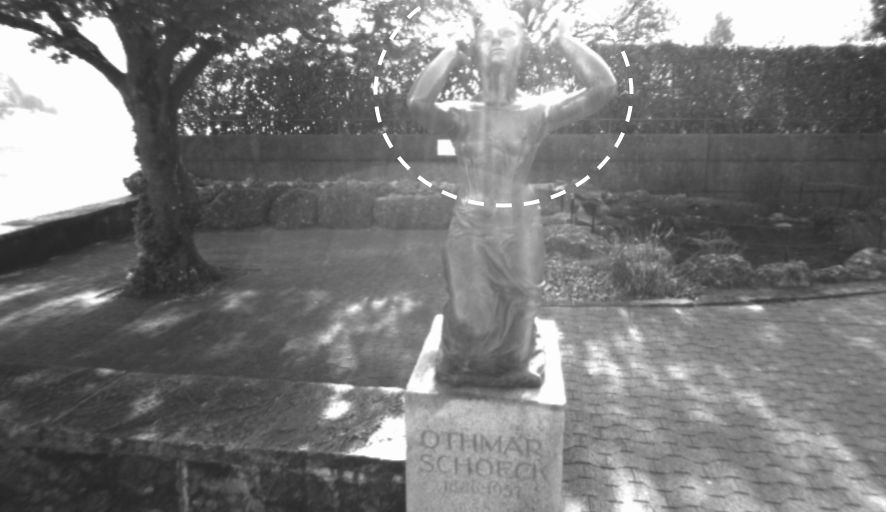}
  \includegraphics[width=.99\linewidth]{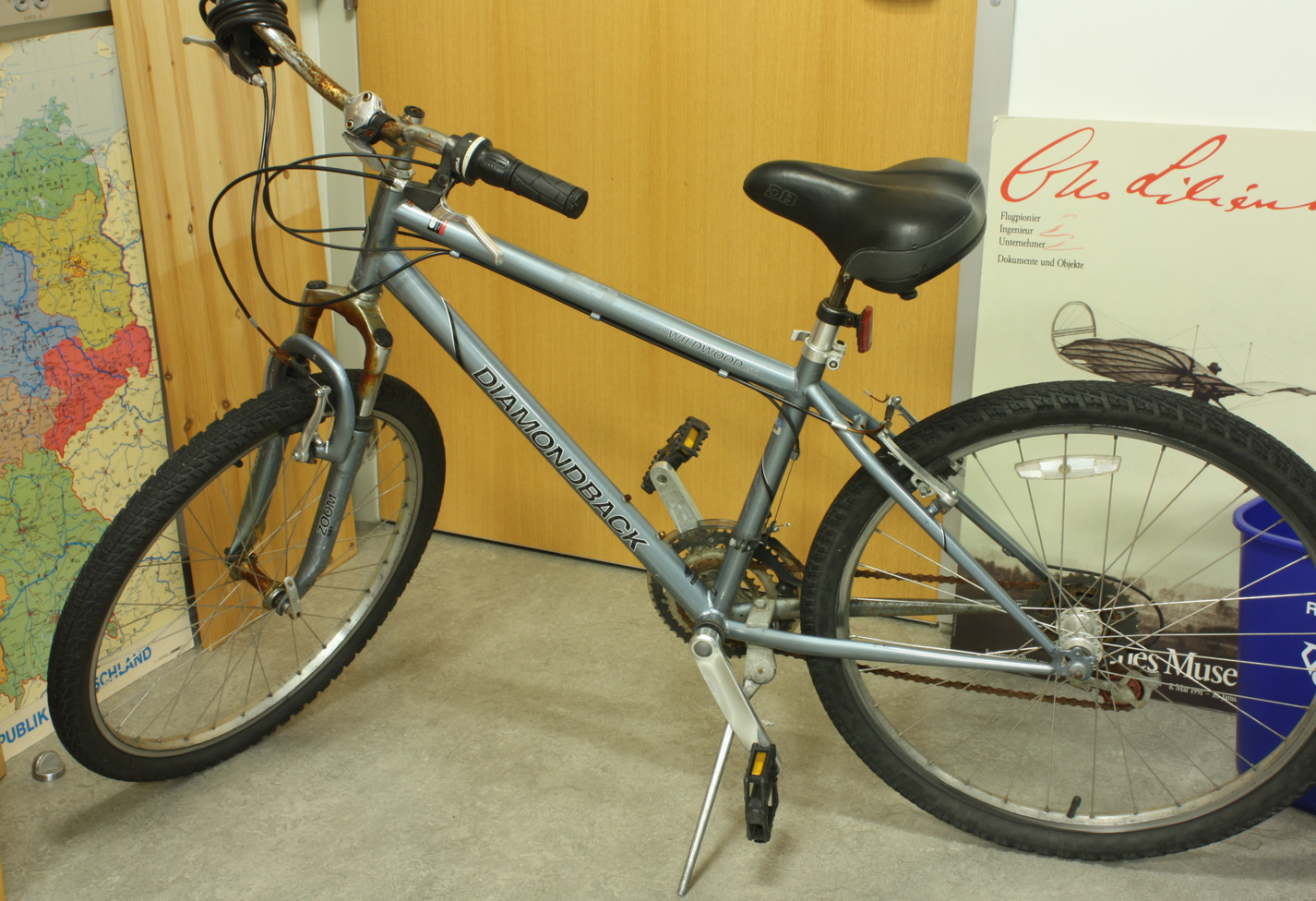}
  \caption{Left Image}
  \label{fig:res_kt12}
\end{subfigure}%
\begin{subfigure}{0.19\linewidth}
  \centering
  \includegraphics[width=.99\linewidth]{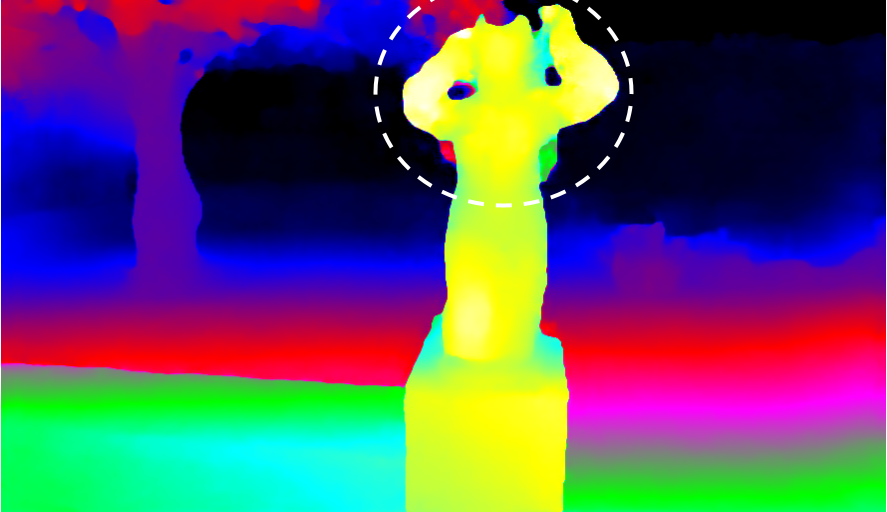}
  \includegraphics[width=.99\linewidth]{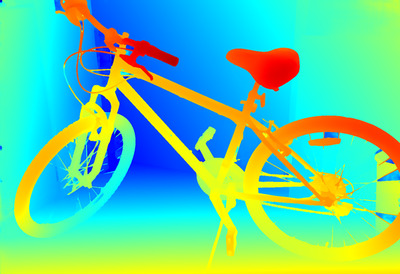}
  \caption{HITNet\cite{tankovich2021hitnet}}
  \label{fig:res_e3d}
\end{subfigure}
% \begin{subfigure}{0.16\linewidth}
%   \centering
%   \includegraphics[width=.99\linewidth]{imgs/lakes1l_dip.png}
%   \caption{DIP-Stereo\cite{zheng2022dip}}
%   \label{fig:res_e3d}
% \end{subfigure}
\begin{subfigure}{0.19\linewidth}
  \centering
  \includegraphics[width=.99\linewidth]{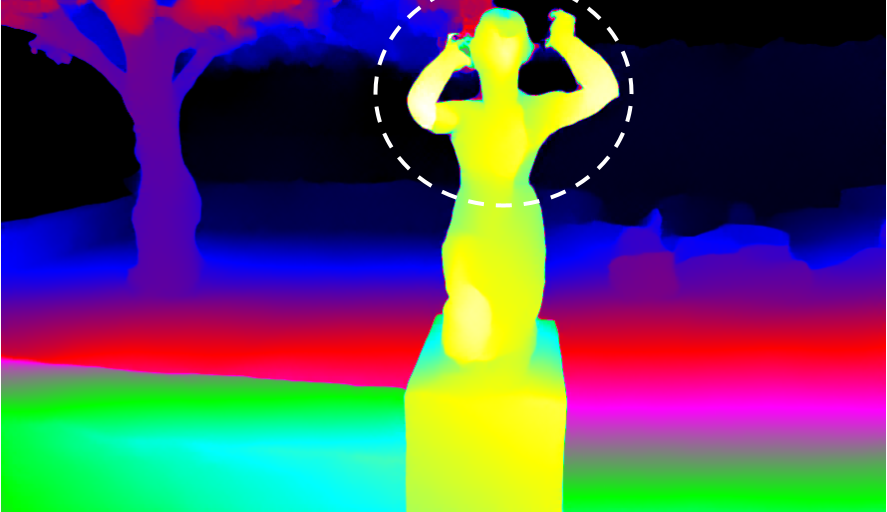}
  \includegraphics[width=.99\linewidth]{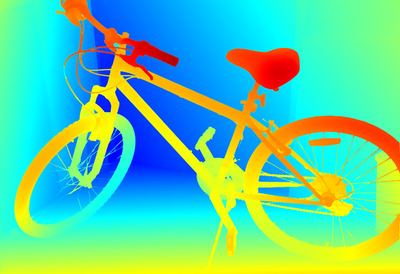}
  \caption{RAFT-Stereo\cite{lipson2021raft}}
  \label{fig:res_e3d}
\end{subfigure}
\begin{subfigure}{0.19\linewidth}
  \centering
  \includegraphics[width=.99\linewidth]{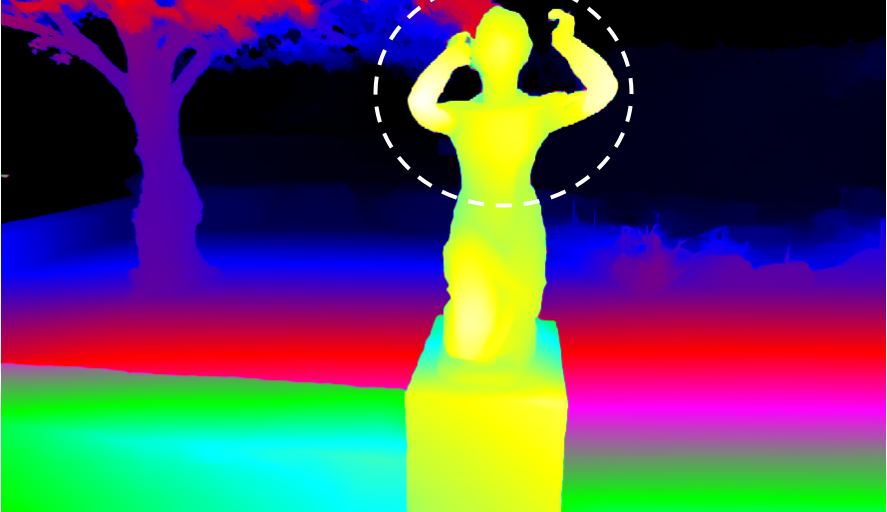}
  \includegraphics[width=.99\linewidth]{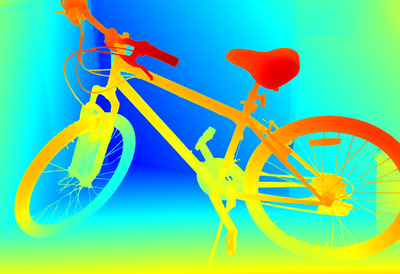}
  \caption{CREStereo\cite{li2022practical}}
  \label{fig:res_e3d}
\end{subfigure}
\begin{subfigure}{0.19\linewidth}
  \centering
  \includegraphics[width=.99\linewidth]{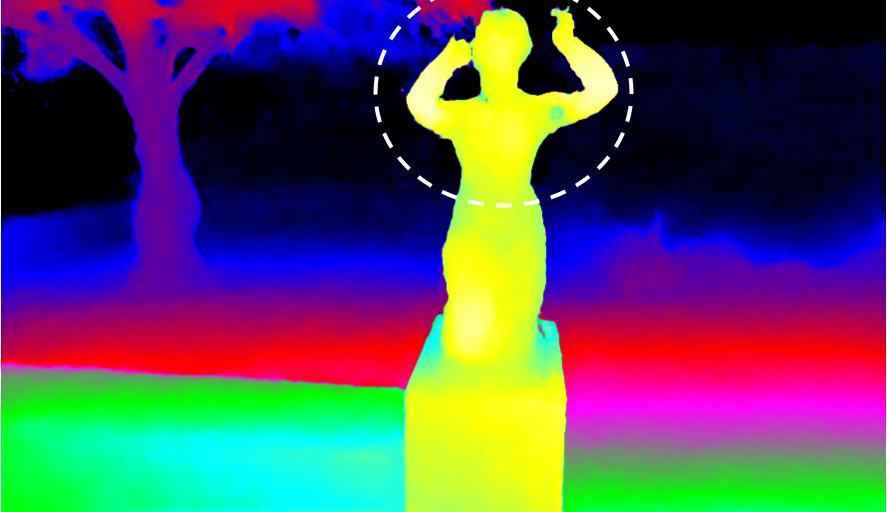}
  \includegraphics[width=.99\linewidth]{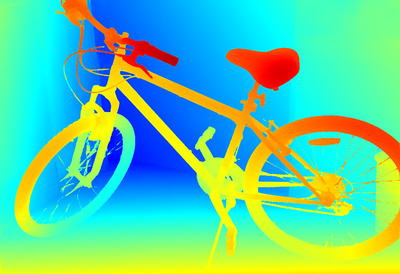}
  \caption{Ours}
  \label{fig:res_e3d}
\end{subfigure}
\caption{
Qualitative comparison on ETH3D benchmark (top) and on Middlebury V3 (bottom).
}
\label{fig:eth3d}
\end{figure*}

\subsection{Ablation Study}\label{sec:ablation}
To evaluate the contributions of different building blocks we test different combinations where group-wise correlation, atrous convolution, and the 2D propagation are enabled/disabled. The counterpart of each building block is: (a) group-wise correlation vs. a single channel in the 3D cost volume constructed by correlation between whole vectors, (b) atrous convolution vs. 2D and 3D convolutions before the broadcasting sum, and (c) keeping or removing the entire 2D propagation.
Each model is trained with the SceneFlow dataset as described in section \ref{sec:training}, and the EPE score is measured on the SceneFlow test set, KITTI 2012, and KITTI 2015.
We also measured runtime and GPU memory consumption during prediction to evaluate the effects of each module on the efficiency.

Results (Table \ref{tab:ablation}) confirm the hypothesis that combining all building blocks leads to the lowest EPE scores. 
Adding the 2D propagation from the backbone, 3D UNet noticeably improved the EPE score (15\%) with small to negligible memory (3.5\%) and runtime (4ms) overhead. 

%Interestingly, on the SceneFlow dataset except for the models only with the 2D pipeline and with the atrous blocks and 2D pipeline.
% In all cases, adding the 2D propagation improves the EPE score with small to negligible memory (2\% to 4\% ) and runtime ($<$7ms) overhead. 
%on the SceneFlow testset at the cost of modest loss on runtime and GPU memory consumption.
%The increase of the GPU memory consumption is from 2\% to 4\% which is affordable for the improvement of the scores, and the runtime loss is less than 7ms which does not harm the overall speed of the model.
% Considering all the results on SceneFlow, KITTI 2012, and KITTI 2015, we selected the model with all modules for other benchmarks.

\subsection{Evaluation}

Similar to the ablation study, we compare our model with other 3D-convolution-based methods on the EPE score, runtime, and GPU memory consumption.
The runtime and GPU memory consumption are measured based on the open-source versions of the respective methods.
As shown in Table \ref{tab:sceneflow}, our model shows the second best EPE score with faster runtime(s) and lower GPU memory consumption among the 3D-convolution-based models. Qualitative comparisons (Figure \ref{fig:sceneflow}) shows that the disparity predicted from our model captures fine-scale features.
To be fair, we want to mention that the current state-of-the-art EPE score (0.36) of the SceneFlow testset is achieved by HITNet\cite{tankovich2021hitnet} based on hierarchical slanted plane refinements and not a 4D cost volume. 

At the same time, we outperform HITNet on the ETH3D benchmark (2nd on Bad 2.0 and Bad 4.0, Table \ref{tab:eth3d}), where we achieve a good trade-off between runtime and accuracy. Figure \ref{fig:eth3d} shows the qualitative comparison for the ``lakeside-1l" sample where our model recovers a clearer silhouette of the statue.
RAFT-Stereo and CREStereo \cite{lipson2021raft, li2022practical} achieve similar to better scores on the ETH3D benchmark, but require several iterations of RNN-based refinements to obtain smooth and clear results. Our model computes disparities in a single run and at runtimes of up to six times faster. 
% utilizing fine-grained initialization of the cost volumes and image warping.

% % ACVNet\cite{xu2022attention} employs GWCNet\cite{guo2019group} as the backbone network which has stacked tripple hourglass network.
% The runtime is

However, at the KITTI 2002 benchmark, we can even outperform RAFT-Stereo and CREStereo with regards to both, runtime and accuracy. The quantitative comparison of state-of-the-art methods on KITTI 2012 is shown in Table \ref{tab:kitti12}. 
On KITTI 2012, our model is ranked 2nd on 3-all, 4-noc, and 4-all but being at least twice as fast as the most recent published state-of-the-art\cite{shenpcw2022pcw} method and up to ten times faster than LaC+GANet which has the closest score\cite{liu2022local}. Qualitative comparisons on KITTI 2012 (Figure \ref{fig:kitti2012}) show again that our model predicts well disparities for fine-scale features.

\section{Conclusion}
In this paper, we proposed image-coupled volume propagation where 2D and 3D propagation collaboratively evolve the cost volumes for stereo matching. This allowed us to design an efficient and accurate model, which ranked 2nd on KITTI 2012 and ETH3D benchmarks while being substantially faster than other comparable methods. Qualitative results demonstrate that our model is capable of capturing detailed structures of the object with the help of 2D propagation.
We also evaluated the building blocks of the model by ablation studies, proofing the efficiency of the accuracy, runtime, and GPU memory consumption. While we used UNet structures as a baseline network, image-coupled volume propagation can be extended to other 3D-convolution-based models regardless of the design. For the future investigations, we see potential especially for methods that combine either real-time framerate or high resolutions with high accuracy and believe that our 2D and 3D propagation may offer interesting avenues to achieve an optimal trade-off.

%%%%%%%%% REFERENCES
\newpage
{\small
\bibliographystyle{ieee_fullname}
\bibliography{main}
}

% \clearpage
% \input{06_appendix}

\end{document}